\documentclass[sigconf]{acmart}
\usepackage{microtype}
\usepackage{graphicx}
\usepackage{subfigure}
\usepackage{booktabs}
\usepackage{colortbl}
\usepackage{tabularx}
\usepackage{multirow}
\usepackage{soul}
\usepackage{balance} 
\usepackage{comment}
\usepackage{tcolorbox}
\definecolor{grey}{RGB}{128,128,128}
\definecolor{LightGray}{gray}{0.9}
\newtcolorbox{infobox}[1][]
{
  colframe = grey!55,
  colback  = grey!10,
  #1
}
\newtheorem{theorem}{Theorem}[section]
\newtheorem{proposition}[theorem]{Proposition}

\theoremstyle{definition}

\usepackage{xspace}
\newcommand{\method}{\text{Hypformer}\xspace}
\usepackage{xcolor}
\newcommand{\darkblue}[1]{{\color{blue!50!black} #1}}
\newcommand{\darkred}[1]{{\color{red!50!black} #1}}
\newcommand{\mdf}[1]{}

\AtBeginDocument{%
  \providecommand\BibTeX{{%
    \normalfont B\kern-0.5em{\scshape i\kern-0.25em b}\kern-0.8em\TeX}}}
\setcopyright{acmlicensed}
\copyrightyear{2024}
\acmYear{2024}
\setcopyright{acmlicensed}\acmConference[KDD '24]{Proceedings of the 30th ACM SIGKDD Conference on Knowledge Discovery and Data Mining}{August 25--29, 2024}{Barcelona, Spain}
\acmBooktitle{Proceedings of the 30th ACM SIGKDD Conference on Knowledge Discovery and Data Mining (KDD '24), August 25--29, 2024, Barcelona, Spain}
\acmDOI{10.1145/3637528.3672039}
\acmISBN{979-8-4007-0490-1/24/08}

\begin{document}

\title{Hypformer: Exploring Efficient Transformer Fully in Hyperbolic Space}

\author{Menglin Yang}
\affiliation{%
  \institution{Yale University}
  \city{New Haven}
  \country{United States}
  mlyang.yale@outlook.com
}
\author{Harshit Verma}
\affiliation{%
  \institution{Birla Institute of Technology and Science}
  \city{Hyderabad}
  \country{India}
  verma08harshit@gmail.com
}
\author{Delvin Ce Zhang}
\affiliation{%
  \institution{Yale University}
  \city{New Haven}
  \country{United States}
  delvincezhang@gmail.com
}
\author{Jiahong Liu}
\affiliation{%
  \institution{The Chinese University of Hong Kong}
  \city{Hong Kong}
  \country{China}
  jiahong.liu21@gmail.com
}
\author{Irwin King}
\affiliation{%
  \institution{The Chinese University of Hong Kong}
  \city{Hong Kong}
  \country{China}
 king@cse.cuhk.edu.hk
}
\author{Rex Ying}
\affiliation{%
  \institution{Yale University}
  \city{New Haven}
  \country{United States}
rex.ying@yale.edu
}
\renewcommand{\shortauthors}{Menglin Yang et al.}
\begin{abstract}
Hyperbolic geometry have shown significant potential in modeling complex structured data, particularly those with underlying tree-like and hierarchical structures. Despite the impressive performance of various hyperbolic neural networks across numerous domains, research on adapting the Transformer to hyperbolic space remains limited.
Previous attempts have mainly focused on modifying self-attention modules in the Transformer.
However, these efforts have fallen short of developing a complete hyperbolic Transformer.
This stems primarily from: (i) the absence of well-defined modules in hyperbolic space, including linear transformation layers, LayerNorm layers, activation functions, dropout operations, etc. (ii) the \textit{quadratic} time complexity of the existing hyperbolic self-attention module w.r.t the number of input tokens, which hinders its scalability. 
To address these challenges, we propose, \textbf{\method}, a novel \textbf{hyp}erbolic Trans\textbf{former} based on the Lorentz model of hyperbolic geometry. In \method, we introduce two foundational blocks that define the essential modules of the Transformer in hyperbolic space. 
Furthermore, we develop a \textit{linear} self-attention mechanism in hyperbolic space, enabling hyperbolic Transformer to process billion-scale graph data and long-sequence inputs for the first time.
Our experimental results confirm the effectiveness and efficiency of \method across various datasets, demonstrating its potential as an effective and scalable solution for large-scale data representation and large models.

\end{abstract}

\ccsdesc[500]{Computing methodologies~Machine learning}
\ccsdesc[500]{Mathematics of computing~Geometric topology}
\ccsdesc[500]{Computing methodologies~Knowledge representation and reasoning}
\keywords{Transformer; Hyperbolic geometry; Linear self-attention; Foundation model}

\maketitle

\section{Introduction}
\label{sec:intro}
\begin{figure}[!t]
\centering
\includegraphics[width=0.35\textwidth]{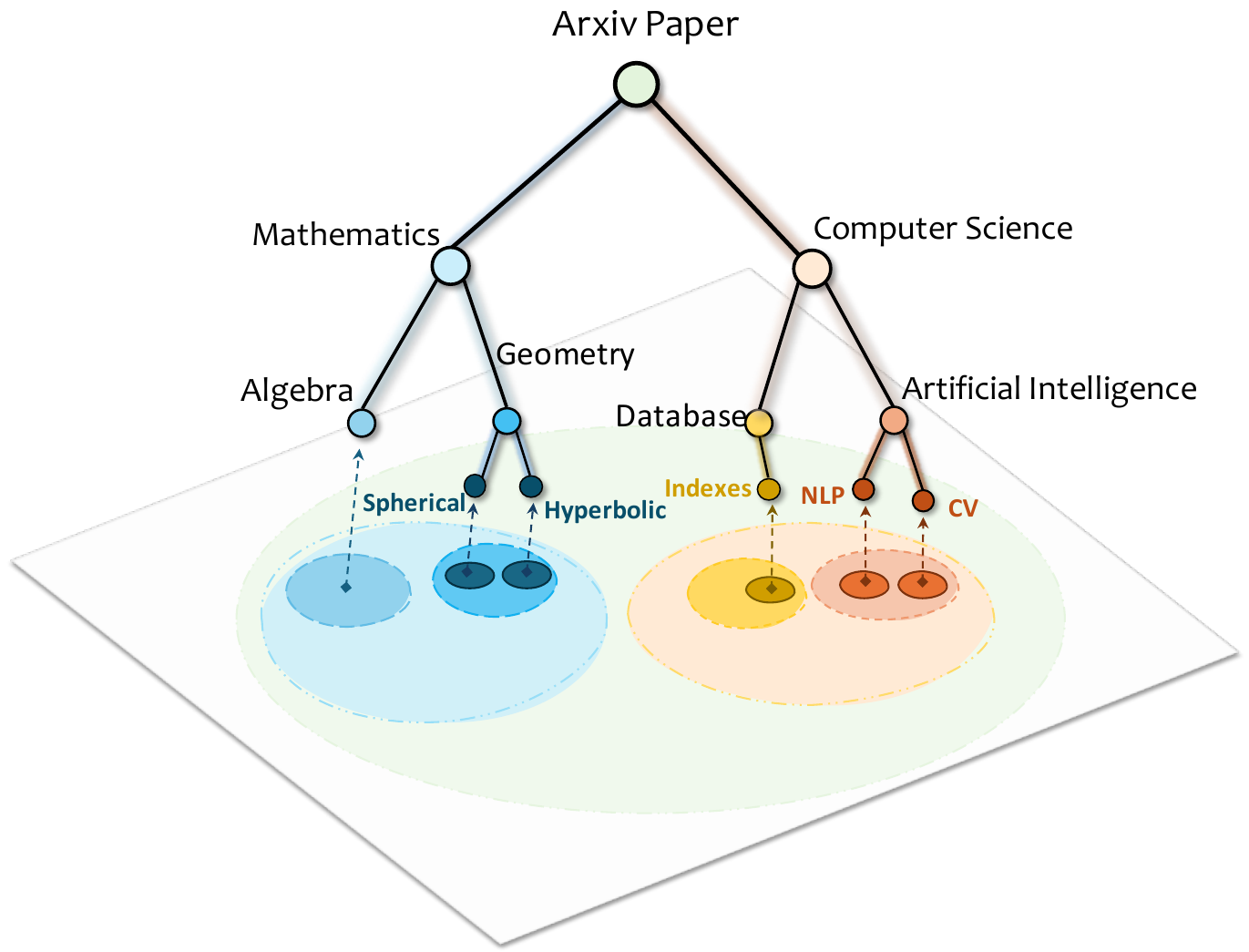}
\caption{In a variety of real-world scenes, when we classify instances in a dataset (e.g., classifying the node in an Arxiv paper), we can group them into larger groups (e.g., Physics, Computer Science) that contain smaller subgroups (e.g., \{Quantum Computing, Geometry\}, \{Database, Artificial Intelligence\}), which may also contain even smaller sub-subgroups. The relationships between these various levels of groups and subgroups can be represented by dendrograms, which are tree-like structures that reveal the underlying hierarchies in the data. }
\label{fig:example}
\vspace{-10pt}
\end{figure}

In many real-world scenarios, data frequently exhibit a hierarchical or tree-like structure, either implicitly or explicitly~\cite{newman2005power,khrulkov2020hyperbolic,zhu2016generative}. 
This is evident in complex networks~\cite{2010hyperbolic,zhou2022telegraph,chami2019hyperbolic,yang2021discrete}, the semantics of words in natural language processing~\cite{nickel2017poincare,nickel2018learning,tifrea2018poincar}, and conceptual hierarchies in vision tasks~\cite{desai2023hyperbolic,khrulkov2020hyperbolic}. As illustrated in Figure~\ref{fig:example}, such data can be organized into large and abstract groups that encompass small and specific subgroups, which can further be subdivided into even smaller and more specific sub-subgroups, and so on. The relationships between these groups and subgroups can be effectively approximated by tree-like structures~\cite{clauset2008hierarchical}. This hierarchical representation mirrors human cognitive processes~\cite{collins1969retrieval,hirtle1985evidence}, making it an intuitive approach to data representation. 

Recent initiatives have explored the use of hyperbolic learning spaces to encode complex non-Euclidean data, achieving impressive performance in representing tree-like data~\cite{nickel2017poincare,nickel2018learning,chami2019hyperbolic,liu2019HGNN,sun2021hgcf,yang2022hrcf,yang2022hicf,HNN,yang2023hyperbolicHIE,yang2022hyperbolic}. 
This success is attributed to the unique property of hyperbolic space, which expands exponentially compared to the polynomial expansion of Euclidean spaces. This property aligns hyperbolic space with the metric of trees, making it particularly suitable for representing tree-like or hierarchically structured data~\cite{2010hyperbolic}.
Despite the growing interest in hyperbolic representation and deep learning, the Transformer~\cite{vaswani2017attention,wang2022foundation,amatriain2023transformer}, a cornerstone model in the various domains, was seldom explored within the realm of hyperbolic space. 
Despite preliminary attempts in hyperbolic Transformers~\cite{gulcehre2019hyperbolicAT,HNN++,chen2021fully}, numerous challenges remain to be addressed.

\textbf{Challenge (1): Insufficient definitions for operations in the hyperbolic Transformer.} 
Prior works of HAN~\cite{gulcehre2019hyperbolicAT} and HNN++~\cite{HNN++} primarily concentrated on the self-attention module, yet they fell short of constructing a comprehensive Transformer architecture, lacking basic components such as LayerNorm layer and positional encoding layer. This is primarily due to the inadequate definition of fundamental operations in previous studies. 

\textbf{Challenge (2): Inefficient and ineffective definitions for linear transformation in the hyperbolic Transformer.}
While some techniques~\cite{HNN,chami2019hyperbolic} employ the tangent space to achieve the linear transformation, they often necessitate frequent logarithmic and exponential mappings, heavily dependent on the tangent space at the origin. This leads to an increased computational load, accumulation of mapping errors, and unstable training procedures. Although \citet{chen2021fully} introduced a fully Lorentz linear transformation in hyperbolic space, it is constrained by its immutable curvature and normalization term.

\textbf{Challenge (3): Absence of a linear attention mechanism in hyperbolic Transformer.} 
The hyperbolic self-attention mechanisms proposed by~\citet{gulcehre2019hyperbolicAT}, \citet{HNN++}, and \citet{chen2021fully} exhibit quadratic time complexity, posing a significant challenge when handling long-sequence input and large-scale graph data.

\textbf{Proposed work:} 
In this work, we propose an efficient hyperbolic Transformer, referred to as \method.
In particular, to address Challenges (1) and (2), we propose two foundational blocks, Hyperbolic Transformation with Curvatures (HTC) and Hyperbolic Readjustment and Refinement with Curvatures (HRC), to build all essential modules in the hyperbolic Transformer. HTC and HRC are built on the Lorentz model of hyperbolic geometry, working directly on the hyperbolic space without frequently mapping.
HTC defines the linear transformation and facilitates mapping from a hyperbolic space with one curvature to another different curvature while preserving the relative distance.
HRC further enables the definition of basic operations commonly used in the Transformer, such as LayerNorm layer, activation function, dropout, and concatenation, within a hyperbolic context. 
To tackle {Challenge (3)}, we introduce a self-attention mechanism in \method with linear complexity, enabling efficient large-scale data processing.

To validate the effectiveness of the proposed methodology, we have undertaken extensive experiments across a diverse range of tasks. These include graph analysis~{\cite{adjacent_encoder,dbn,li2022bsal,yang2020featurenorm,liu2022discovering}}, text classification~{\cite{hgtm,tag}}, and image classification~{\cite{dhall2020hierarchical,wu2022nodeformer}}. The empirical evidence gathered from these experiments indicates that the proposed method significantly reduces the GPU computation cost by a factor of 10 and concurrently halves the training time compared with the existing hyperbolic softmax attention. Furthermore, the proposed method consistently surpasses the performance of competitive baselines, yielding substantial improvements on both tree-like and non-tree-like datasets.

\textbf{Contributions.} In summary, this study offers the following contributions: 
\textit{First}, we introduce two fundamental hyperbolic blocks, HTC and HRC. Building upon these, we have formulated fundamental modules for linear transformation, LayerNorm, activation function, dropout, and concatenation operations within a hyperbolic context. 
\textit{Second}, we propose the first hyperbolic linear attention mechanism, which enables the hyperbolic Transformer to be scalable and efficient. Based on the above efforts,
we construct a \method\footnote{\darkblue{{Code is available at \url{https://github.com/Graph-and-Geometric-Learning/hyperbolic-transformer}}}}, the first comprehensive and efficient hyperbolic Transformer model fully designed to operate within hyperbolic space.
\textit{Last}, we extend the hyperbolic model to handle billion-level graph data for the first time, laying a crucial foundation for the application of big data and large-scale models.

\section{Related Work}
\subsection{Hyperbolic Neural Networks}
\label{sec:related_work} 
Recent studies have demonstrated that hyperbolic space is particularly adept at capturing the hierarchical and tree-like structures~\cite{nickel2017poincare,nickel2018learning,ganea2018hyperbolic,law2019lorentzian,gu2019learning,tifrea2018poincar,yang2023hyperbolicHIE,liu2022enhancing,Qiu00K24}. 
Building on hyperbolic space, a variety of hyperbolic neural networks, HNN~\cite{HNN}, HAN~\cite{gulcehre2019hyperbolicAT}, HNN++~\cite{HNN++}, HGCN~\cite{chami2019hyperbolic}, HGNN~\cite{liu2019HGNN}, F-HNN~\cite{chen2021fully}, Poincar\'e Resnet~\cite{van2023poincar}, HGTM~\cite{hgtm} have been developed to leverage the advantages of the hyperbolic geometry. 
These neural networks have obtained an impressive performance in domains like computer vision~\cite{khrulkov2020hyperbolic,atigh2022hyperbolic,hsu2021capturing}, natural language processing~\cite{montella2021hyperbolic,kolyvakis2019hyperkg,bai2021modeling,chami2020low}, recommender systems~\cite{HyperML2020,yang2022hrcf,sun2021hgcf,wang2021hypersorec,yang2022hicf,chen2022modeling}, graph learning~\cite{zhang2021hyperbolic,chami2019hyperbolic,liu2019HGNN,liu2022enhancing,yang2021discrete,bai2023hgwavenet,yang2023kappahgcn} and so on~\cite{xiong2022hyperbolic,liu2022enhancing}.

\subsection{Transformer and Hyperbolic Transformer} 

Introduced by \citet{vaswani2017attention}, Transformer models have brought about a paradigm shift in the field of artificial intelligence. 
Transformer \cite{vaswani2017attention, devlin2018bert, brown2020language, dosovitskiy2020image} has made a tremendous impact in many fields, such as language understanding~\cite{devlin2018bert,brown2020language, raffel2020exploring}, image processing~\cite{parmar2018image, carion2020end} and graph learning~\cite{kim2022pure,rampavsek2022recipe}.
A well-known concern with self-attention is the quadratic time complexity, which can hinder model scalability in many settings. Efficient self-attention models are crucial in applications that model long sequences~\cite{kitaev2020reformer,roy2021efficient,katharopoulos2020transformers,han2023flatten}.

Despite these advancements, existing Transformer architectures predominantly operate within the Euclidean domain. There have been limited attempts to extend these models to hyperbolic and other non-Euclidean spaces. \citet{gulcehre2019hyperbolicAT} proposed hyperbolic attention networks, which replace the dot-product between the \textit{query} and \textit{key} in self-attention with a function of negative hyperbolic distance. 
They then utilize the Einstein midpoint to compute the attentive output with \textit{value}. Similarly, \citet{chen2021fully} and \citet{HNN++} adopt similar strategies that result in the attentive output with \textit{key} being based on the Lorentzian midpoint and gyromidpoint, respectively.\footnote{In theory, the Einstein midpoint, Lorentzian centroid, and gyromidpoint are equivalent midpoint operations projected onto each manifold~\cite{HNN++}.} 
However, these methods exhibit quadratic time complexity, limiting their scalability. Besides, they focused more on the self-attention module and did not define the essential modules, like LayerNorm in Transformer. 
Recently, \citet{cho2023curve} proposed a fully Product-Stereographic Transformer, presenting a kernelized approach to non-Euclidean attention, which is linear time complexity.
However, this method heavily relies on the tangent space, necessitating frequent mappings between the tangent space and manifolds. 
\citet{ermolov2022hyperbolic} proposed mapping the last layer features obtained from a Euclidean Transformer to hyperbolic space, which essentially does not establish a true Hyperbolic Transformer.
Our work aims to address these challenges and further the development of hyperbolic Transformers.

\section{Preliminaries}
\label{sec:preliminaries}
In this section, we introduce concepts related to Lorentz model of hyperbolic geometry and self-attention module briefly.

\subsection{Lorentz Model of Hyperbolic Geometry}
There are several isometric models~\cite{nickel2017poincare,ganea2018hyperbolic,tifrea2018poincar,gulcehre2019hyperbolicAT,nickel2018learning,ramsay2013introduction} of hyperbolic geometry that have been employed in prior research. 
In this study, we choose the Lorentz model as the foundational framework due to the numerical stability it offers~\cite{nickel2018learning,mishne2023numerical}. Also, the proposed \method can be easily adapted to other hyperbolic models, as they are isometrically equivalent.

\textbf{Lorentz Model.} 
An $n$-dimensional Lorentz model with negative constant curvature $\kappa (\kappa<0)$ is a Riemannian manifold denoted by $\mathbb{L}^{n,\kappa}$. The corresponding Riemannian metric is given by $\mathfrak{g}^\kappa=\operatorname{diag}(1/k,1, \cdots, 1)$. 
Each point in $\mathbb{L}^{n,\kappa}$ can be represented as $\mathbf{x}=\left[\begin{array}{c}x_t \\ \mathbf{x}_s\end{array}\right]$ 
where $\mathbf{x} \in \mathbb{R}^{n+1}$, $x_t \in \mathbb{R}$ and $\mathbf{x}_s \in \mathbb{R}^n$. The set of points, $\mathbb{L}^{n,k}$, that constitute the manifold are defined as 
\begin{equation}
\mathbb{L}^{n, \kappa}:=\left\{\mathbf{x} \in \mathbb{R}^{n+1} \mid \langle\mathbf{x}, \mathbf{x}\rangle_{\mathcal{L}}= 1/{\kappa}, x_t>0\right\}.  
\label{equ:lorentz_defination}
\end{equation}
Here, $\langle\mathbf{x}, \mathbf{y}\rangle_{\mathcal{L}}=-x_t y_t+\mathbf{x}_s^{\top} \mathbf{y}_s=\mathbf{x}^{\top} \mathfrak{g}^\kappa \mathbf{y}$ represents the Lorentzian inner product. 
Lorentz model, also known as the hyperboloid model, is an upper hyper-surface in an $(n+1)$ dimensional Minkowski space with the origin point $(\sqrt{-1/\kappa}, 0, \cdots, 0)$. Lorentz model has its roots in the theory of special relativity~\cite{resnick1991introduction} and employs terminology borrowed from this field. The hyperboloid's axis of symmetry, represented by the 0-th element $x_t$, is referred to as the time-like dimension, while all other axes $\mathbf{x}_s$ are called space-like dimensions.

\textbf{Tangent Space of Lorentz Model.} Given $\mathbf{x} \in \mathbb{L}^{n,\kappa}$, the tangent space $\mathcal{T}_{\mathbf{x}} \mathbb{L}^{n,\kappa}:=\left\{\mathbf{u} \in \mathbb{R}^{n+1} \mid\langle\mathbf{u}, \mathbf{x}\rangle_{\mathcal{L}}=0\right\}$ is the orthogonal space of $\mathbb{L}^{n,\kappa}$ at $\mathbf{x}$ with respect to the Lorentzian inner product.\footnote{The orthogonality condition \(\langle \mathbf{u}, \mathbf{x} \rangle_{\mathcal{L}} = 0\) ensures that \(\mathbf{u}\) lies in the tangent space, preserving the manifold's geometry.}
To achieve the mapping from the Lorentz model to the tangent space at $\mathbf{x}$, we can use the logarithmic map, $\log_\mathbf{x}^\kappa:\mathbb{L}^{n,\kappa}\to \mathcal{T}_{\mathbf{x}} \mathbb{L}^{n,\kappa}$. The exponential map defines the inverse process, $\exp_\mathbf{x}^\kappa: \mathcal{T}_{\mathbf{x}} \mathbb{L}^{n,\kappa}\to \mathbb{L}^{n,\kappa}$. For the details about exponential, logarithmic maps and the relevant distance functions, please refer to Appendix~\ref{appendix:exponentail_and_logrithmic_map}.

\begin{figure}[!t]
\centering
\includegraphics[width=0.28\textwidth]{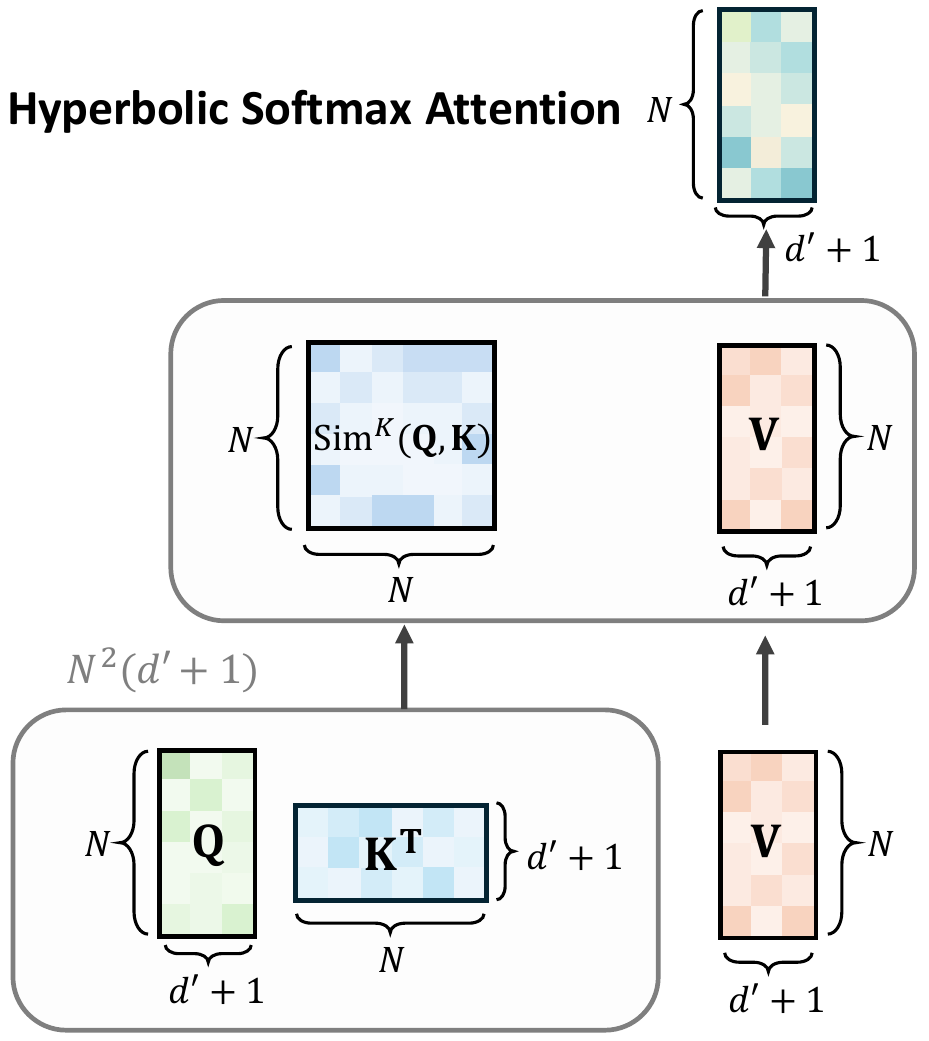}
\caption{Illustration of hyperbolic softmax attention defined on Lorentz model. Unlike the attention mechanism in Euclidean space, this hyperbolic attention obtains the similarity between $\mathbf{Q}$ and $\mathbf{K}$ by a negative hyperbolic distance defined in Equation~(\ref{equ:hyperbolic_sim_k}). The time complexity is quadratic w.r.t the number of input tokens.
}
\label{fig:softmax_attention}
\end{figure}

\subsection{Self-Attention Module}
We first examine the general form of self-attention in Euclidean Transformers. Given the input of $\mathrm{N}$ tokens $\mathbf{X} \in \mathbb{R}^{N \times d}$, within each head, self-attention can be expressed as:
\begin{equation}
\begin{gathered}
\mathbf{Q}= \mathbf{X}\mathbf{W}^\mathbf{Q}, \mathbf{K}=\mathbf{X}\mathbf{W}^\mathbf{K}, \mathbf{V}=\mathbf{X}\mathbf{W}^\mathbf{V}, \\
\mathbf{Z}_i=\sum_{j=1}^N \frac{\operatorname{Sim}\left(\mathbf{Q}_i, \mathbf{K}_j\right)}{\sum_{j=1}^N \operatorname{Sim}\left(\mathbf{Q}_i, \mathbf{K}_j\right)} \mathbf{V}_j,
\end{gathered}
\label{equ:self-attention}    
\end{equation}
where $\mathbf{W}^\mathbf{Q}, \mathbf{W}^\mathbf{K}, \mathbf{W}^\mathbf{V} \in \mathbb{R}^{d \times d'}$ are projection matrices and $\operatorname{Sim}(\cdot, \cdot)$ denotes the similarity function. Modern Euclidean Transformers primarily use Softmax attention~\cite{vaswani2017attention} where similarity is calculated as $\operatorname{Sim}(\mathbf{Q}_i, \mathbf{K}_j)=\exp \left(\mathbf{Q}_i \mathbf{K}_j^T / \sqrt{d'}\right)$. In this scenario, the attention map is derived by computing the similarity between all query-key pairs, which results in a computational complexity of $\mathcal{O}\left(N^2\right)$.

The concept of hyperbolic self-attention, as defined by previous works~\cite{gulcehre2019hyperbolicAT,HNN++,chen2021fully}, bears a similar idea to Equation~(\ref{equ:self-attention}). Figure~\ref{fig:softmax_attention} presents an illustration for this hyperbolic operation on Lorentz model. It can be expressed as follows:
\begin{equation}
\begin{gathered}
\mathbf{Q}= \mathbf{X}\otimes^\kappa\mathbf{W}^\mathbf{Q}, \mathbf{K}=\mathbf{X}\otimes^\kappa\mathbf{W}^\mathbf{K}, \mathbf{V}=\mathbf{X}\otimes^\kappa\mathbf{W}^\mathbf{V}, \\
\mathbf{Z}_i=\sum_{j=1}^N \frac{\operatorname{Sim}^\kappa\left(\mathbf{Q}_i, \mathbf{K}_j\right)}{\sum_{j=1}^N \operatorname{Sim}^\kappa\left(\mathbf{Q}_i, \mathbf{K}_j\right)} \odot^\kappa\mathbf{V}_j.
\end{gathered}
\label{equ:self-attention_lorentz}    
\end{equation}
In this equation, $\otimes^\kappa$ denotes the hyperbolic linear transformation, which can be computed using Equations (\ref{equ:tangent_linear_transforamtion}) and (\ref{equ:hyperbolic_linear_transforamtion}) given in the following section. The symbol $\odot^\kappa$ represents the weighted sum in hyperbolic space. Let $\mathbf{Att}_i$ denotes the $i$-th row of the attention matrix in the Lorentz model, it can be computed by Lorentzian midpoint~\cite{law2019lorentzian}:
\begin{equation}
\begin{aligned}
\mathbf{Att}_i\odot^\kappa\mathbf{V}_j:=&\frac{\sum_{j=1}^{{N}} \alpha_{i j} \mathbf{V}_j}{\sqrt{|{\kappa}|\left|\|\sum_{k=1}^{{N}} \alpha_{i k} \mathbf{V}_k \|_{\mathcal{L}}\right|}}. \\
\end{aligned}
\end{equation}
The function $\operatorname{Sim}^\kappa(\cdot, \cdot)$ denotes the similarity function defined by the hyperbolic distance $d_\mathcal{H}$~\cite{gulcehre2019hyperbolicAT,HNN++,chen2021fully}\footnote{Here we use subscript $\mathcal{H}$ other than $\mathcal{L}$ since it is not limited to Lorentz mdoel.} or the tangent inner product~\cite{micic2018hyperbolic}.
Specifically, \citet{chen2021fully} defined the similarity function as:
\begin{equation}
    \operatorname{Sim}^\kappa(\mathbf{Q}_i, \mathbf{K}_j)=\exp \left(-d_\mathcal{H}^2(\mathbf{Q}_i, \mathbf{K}_j) / \sqrt{d'}\right).
    \label{equ:hyperbolic_sim_k}
\end{equation}
Both \citet{gulcehre2019hyperbolicAT} and \citet{HNN++} utilized similar forms of this function. They all employ negative distance to define similarity, and each has a computational complexity of $\mathcal{O}\left(N^2\right)$.

\subsection{Lorentz Transformation}
\textbf{Lorentz Tangent Space Transformation}. Previous works~\cite{chami2019hyperbolic,liu2019HGNN,yang2022htgn,HNN,zhang2021lorentzian} mainly define hyperbolic linear transformations by the tangent space method, termed as $\mathrm{LT}_\mathcal{T}$. Given the Lorentz embedding vector $\mathbf{x}$ and operation function $f$, the tangent space method maps $\mathbf{x}$ to the tangent space at a local reference point by the logarithmic map. Then, the transformation operation $f$ is applied in this tangent space. Finally, the resulting vector is mapped back to the Lorentz model using the exponential mapping, that is\footnote{Some studies~\cite{zhang2021lorentzian, yang2022htgn} proposed an improved version of tangential linear transformations only on the space-like dimension and then incorporated a zero value to the transformed results, in order to respect the constraints of the tangent space at the origin. They have a similar formula as Equation~(\ref{equ:tangent_linear_transforamtion}), which we omit for brevity.}, 
\begin{equation}
\mathrm{LT}_\mathcal{T}{(\mathbf{x}; f, \kappa_1, \kappa_2)} := \exp_{{\mathbf{o}}}^{\kappa_1}(f(\log_{{\mathbf{o}}}^{\kappa_2}(\mathbf{x}))),
\label{equ:tangent_linear_transforamtion}
\end{equation}
where $\mathbf{o}$ is the local reference point (generally the origin point), and the curvatures $\kappa_1$ and $\kappa_2$ could be different since they share the same tangent space. Using this method, previous works define the linear transformation, neighbor's aggregation, dropout, and non-linear activation~\cite{chami2019hyperbolic,liu2019HGNN}.

\textbf{Limitations}. While this method is intuitive, it has notable limitations. \textit{First}, parallel computation is feasible if the same reference point is used for the entire embedding. However, this approach can lead to significant mapping errors for distant points due to the point-specific nature of the tangent space. Conversely, using local-specific points enhances accuracy but increases computational load by requiring separate mappings. \textit{Second}, frequent use of hyperbolic functions like cosh or cosh$^{-1}$ can destabilize learning. While the clamp function can mitigate this issue, its use may compromise computational precision.

\textbf{Fully Lorentz Transformation}. To overcome the above limitations, \citeauthor{chen2021fully}~\cite{chen2021fully} defined an alternative Lorentz transformation without using tangent space, termed as $\mathrm{LT}_\mathcal{F}$:
\begin{equation}
\mathrm{LT}_\mathcal{F}(\mathbf{x};f, \mathbf{W}, \kappa) := \left({\sqrt{\|f(\mathbf{W x}, \mathbf{v})\|^2-1 / \kappa}}, {f(\mathbf{W x}, \mathbf{v})}\right)^T,
\label{equ:hyperbolic_linear_transforamtion}
\end{equation}
which involves a function $f$ that operates on vectors $\mathbf{v} \in \mathbb{R}^{n+1}$ and $\mathbf{W} \in \mathbb{R}^{m \times(n+1)}$, $f(\mathbf{W x}, \mathbf{v})=$ $\frac{\lambda \sigma\left(\mathbf{v}^{\top} \mathbf{x}+b^{\prime}\right)}{\|\mathbf{W} h(\mathbf{x})+\mathbf{b}\|}(\mathbf{W} h(\mathbf{x})+\mathbf{b})$. Here, $\sigma$ is the sigmoid function, $\mathbf{b}$ and $b^{\prime}$ are bias terms, $\lambda>0$ controls the scaling range, and $h$ is the activation function. Depending on the type of function, it can perform different operations. For instance, for dropout, the operation function is $f(\mathbf{W} \mathbf{x}, \mathbf{v})=\mathbf{W}$ dropout $(\mathbf{x})$. 

\textbf{Limitations}. There are several limitations to this method. 
\textbf{First}, the curvature is unchangeable. Although it appears that $\mathrm{LT}_\mathcal{F}$ provides a way to directly modify $\kappa$ in Equation~(\ref{equ:hyperbolic_linear_transforamtion}), this modification results in a loss of previously learned information, introducing distortions. \textit{Direct alteration of curvature cannot guarantee the preservation of relative distance relationships within the learned embedding.}
The derivation is shown as follows:
\begin{infobox}[]
Let $\mathbf{x}^{\prime} = f(\mathbf{W}\mathbf{x}, \mathbf{v})$, and $g(\mathbf{x}^{\prime}) = \left(\sqrt{\|\mathbf{x}^{\prime}\|^2 - 1/\kappa^{\prime}}, \mathbf{x}^{\prime}\right)$. Then:
\begin{equation}
\begin{aligned}
d_{\mathcal{L}}^{\kappa^{\prime}}\left(g(\mathbf{x}^{\prime}), g(\mathbf{y}^{\prime})\right) &= \sqrt{1/{|\kappa|}^{\prime}}\text{arcosh}\left(\kappa'\langle g(\mathbf{x}^{\prime}), g(\mathbf{y}^{\prime}) \rangle_\mathcal{L}\right)\\
    &= \sqrt{1/{|\kappa|}^{\prime}}\text{arcosh}\left( \kappa^{\prime}\left(\alpha_\text{time} + \mathbf{x}^{{\prime}T}\mathbf{y}^{\prime} \right)\right),
\end{aligned}
\label{equ:distance}
\end{equation}
where $\alpha_\text{time} = \sqrt{\left(\|\mathbf{x}^{\prime}\|^2 - 1/ \kappa^{\prime}\right)\left(\|\mathbf{y}^{\prime}\|^2 - 1/ \kappa^{\prime}\right)}$. 
\end{infobox}
It can be observed that changing $\kappa$ results in a non-linear transformation of the Lorentz distance $d_{\mathcal{L}}^{\kappa^{\prime}}$. Consequently, the relative distances between data points may not be preserved as they were in the original $\kappa$ Lorentz space. Even small changes in the parameter $\kappa$ can significantly affect the resulting distances, potentially distorting the previously learned hierarchical structure.

\textbf{Second}, the requirement for the $\mathbf{W}$ matrix and normalization term pose another challenge. In~\cite{chen2021fully}, $\mathbf{W}$ is applied to both time-like and space-like dimensions , in order to achieve Lorentz boosts and rotations simultaneously. However, its introduction constrains the usage of certain functions. For instance, dropout, activation operation do not necessarily interact with the matrix $\mathbf{W}$. Taking the ReLU activation function as an example, it only requires filtering out negative values without needing matrix multiplication in Euclidean space. Additionally, \citet{chen2021fully} introduced a normalization term that constrains the value within a limited range, thereby limiting the expressiveness of the transformation.

\textbf{Lastly}, some basic operations, such as LayerNorm and Concatenation, cannot be achieved within this definition.

\begin{figure}[!t]
\centering
\includegraphics[width=0.28\textwidth]{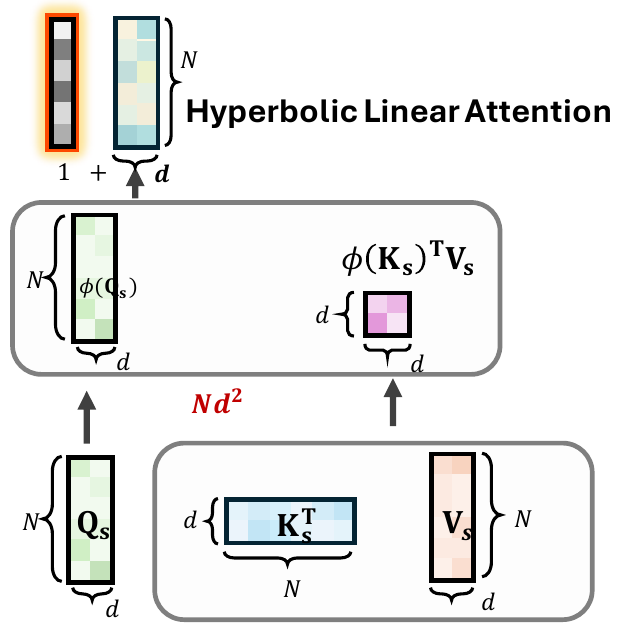}
\caption{Illustration of Hyperbolic Linear attention. This attention operates in the space-like dimension ($\mathbf{Q}_s$, $\mathbf{K}_s$, $\mathbf{V}_s$) and reduces the time complexity by changing the computation order.}
\label{fig:linear_attention}
\vspace{-20pt}
\end{figure}
\section{Method}
\label{sec:method}
The proposed method is designed to overcome the limitations of the existing attempts in hyperbolic Transformer, as outlined in the Section~\ref{sec:intro}. To address Challenges (1) and (2), we designed two foundational blocks, namely HTC and HRC in Section \ref{sec:htc}, and \ref{sec:hrc}, respectively.  
To overcome Challenge (3), we developed a hyperbolic linear attention module in Section~\ref{sec:linear_attention}, which equips the Transformer with linear time complexity.

\subsection{Hyperbolic Transformation with Curvatures (HTC)}
\label{sec:htc}
\textbf{Novelty.} Unlike the design of linear transformation using the tangent space method in Equation~(\ref{equ:tangent_linear_transforamtion}), we build the transformation fully in hyperbolic space. 
Besides, compared with Lorentz transformation defined by Equation~(\ref{equ:hyperbolic_linear_transforamtion}), we have two improvements: (1) making the curvature changeable with preserving the relative ordering; (2) being disentangled with normalization term. 

Given a point $\mathbf{x}$ in Lorentz model, \( \mathbf{x} \in \mathbb{L}^{d,\kappa_1} (\mathrm{implies}~\mathbf{x} \in \mathbb{R}^{d+1}) \), and transformation matrix $\mathbf{W}\in \mathbb{R}^{{(d+1)}\times d'}$ and bias ${b}\in \mathbb{R}^{d'}$, the HTC is given as the following equation:
\begin{equation}
\begin{aligned}
        \operatorname{HTC}(\mathbf{x}; f_t, \mathbf{W},\kappa_1, \kappa_2):=\left(\underbrace{\sqrt{\frac{\kappa_1}{\kappa_2}\| f_t(\mathbf{x}; \mathbf{W}) \|_2^2-1/\kappa_2}}_{\text {time-like dimension }}, \underbrace{\sqrt{\frac{\kappa_1}{\kappa_2}} f_t(\mathbf{x}; \mathbf{W})}_{\text {space-like dimension }}\right)^T,
\end{aligned}
\end{equation}
where the $f_t(\mathbf{x}; \mathbf{W}) = \mathbf{W}^T\mathbf{x} + {b}$ denotes the linear transformation with bias addition and $\kappa_1$, $\kappa_2$ represent the curvatures before and after the transformation. Note that HTC does not entangle a normalization term with this linear transformation. \textbf{Besides,
It is easy to prove that the defined transformation also satisfies the Lorentz rotation and boost operations, described in~\cite{chen2021fully}.} The proof is similar in ~\cite{chen2021fully}, we omit for brevity. 

The proposed HTC avoids the use of tangent space and minimizes the usage of logarithmic and exponential mappings in comparison to Equation~(\ref{equ:tangent_linear_transforamtion}). When contrasted with Equation (\ref{equ:hyperbolic_linear_transforamtion}), the variable curvature of the HTC enhances the flexibility of the transformation. This is because linear transformations generally alter the feature dimension, and varying curvatures can express more than a fixed one. 

Next, we study the theoretical aspects of the proposed HTC. First and foremost, we prove that HTC is closed in hyperbolic space in Proposition~(\ref{prop:lorentz_constraint}) with different dimensions and curvatures so that the mapping is done correctly. Next, in Proposition~(\ref{prop:lorentz_preserving_text}),  we show that the curvature-changing strategy of the proposed HTC, along with the subsequent HRC, maintains the relative distance among any points between pre and post-curvature changing.

\begin{proposition}
Let $\mathbf{x} \in \mathbb{L}^{d_a, \kappa_a}$ and $\mathbf{W} \in \mathbb{R}^{(d_a + 1) \times d_b}$. The LTC operation, defined as $\operatorname{LTC}(\mathbf{x}; f_t; \mathbf{W}, \kappa_a, \kappa_b)$, correctly transforms $\mathbf{x}$ from the Lorentz model with curvature $\kappa_a$ to the Lorentz model with curvature $\kappa_b$, such that
\begin{equation}
\operatorname{LTC}(\mathbf{x}; f_t; \mathbf{W}, \kappa_a, \kappa_b) \in \mathbb{L}^{d_b, \kappa_b}.
\end{equation}
\label{prop:lorentz_constraint}
\end{proposition}

\begin{proposition}
Let $\mathbf{z}_i, \mathbf{z}_j, \mathbf{z}_k \in \mathbb{L}^{\kappa_a}$ be points in the Lorentz model with curvature $\kappa_a$. Consider the curvature changing transformations defined in HTC (Equation~(\ref{equ:hyperbolic_linear_transforamtion})) and HRC (Equation~\ref{equ:hrc}). Let $\mathbf{z}_i', \mathbf{z}_j', \mathbf{z}_k' \in \mathbb{L}^{\kappa_b}$ denote the transformed points in the Lorentz model with curvature $\kappa_b$.
The relative distances within ($\mathbf{z}_i, \mathbf{z}_j, \mathbf{z}_k$) are preserved after the curvature alteration. Specifically, if
\begin{equation}
d_\mathcal{L}^{\kappa_a}(\mathbf{z}_i, \mathbf{z}_j) \geq d_\mathcal{L}^{\kappa_a}(\mathbf{z}_i, \mathbf{z}_k),
\end{equation}
then
\begin{equation}
d_\mathcal{L}^{\kappa_b}(\mathbf{z}_i', \mathbf{z}_j') \geq d_\mathcal{L}^{\kappa_b}(\mathbf{z}_i', \mathbf{z}_k').
\end{equation}
\label{prop:lorentz_preserving_text}
\end{proposition}

\subsection{Hyperbolic Readjustment and Refinement with Curvatures (HRC)}
\label{sec:hrc}
\textbf{Novelty.} Within the Transformer, we have several basic operations beyond linear transformation, which include \textit{Dropout} and \textit{Concatenation}, \textit{Activation function} (e.g., ReLU), and \textit{LayerNorm}. 
We interpret these operations within the hyperbolic space as a readjustment or refinement process, referred to as HRC. 
Similarly, given a point $\mathbf{x}$ in Lorentz model, the proposed operation $\mathrm{HRC}$ is defined as:
\begin{equation}
\begin{aligned}
    \mathrm{HRC}(\mathbf{x};f_r, \kappa_1, \kappa_2) := \left( \underbrace{\sqrt{\frac{\kappa_1}{\kappa_2} \|f_r(\mathbf{x}_{[1:]})\|_2^2 - 1/\kappa_2}}_{\text{time-like dimension}}, \underbrace{\sqrt{\frac{\kappa_1}{\kappa_2}}f_r(\mathbf{x}_{[1:]})}_{\text{space-like dimension}} \right)^T.
\end{aligned}
\label{equ:hrc}
\end{equation}
Here, $f_r$ represents a function applied to the space-like dimensions. 

It is evident that HRC shares similar advantages with HTC, which we will not repeat for the sake of brevity. However, unlike HTC, HRC performs the transformation only in space-like dimensions. The primary motivation is as follows:
HTC involves a Lorentz boost, essential for mapping between \textbf{different} inertial reference frames, tied to causality and affecting the observed sequence of events in relativity. 
However, operations such as LayerNorm, activation functions, dropout, and concatenation serve as readjustments or refinements within the \textbf{same} frame of reference, acting on space-like features to standardize, activate, or regularize them. Applying these to the space-like dimension ensures the causal structure remains intact. In practical, it ensures dimensional consistency, improves interpretability, and allows for more efficient computation.
Nonetheless, it is important to note that HRC does not completely discard the time-like information. According to the definition in Equation~(\ref{equ:lorentz_defination}), the time-like dimension is determined by the space-like dimensions. By operating on the space-like dimensions, HRC implicitly utilizes the time-like information.

\begin{figure*}[h]
\centering
\includegraphics[width=0.95\textwidth]{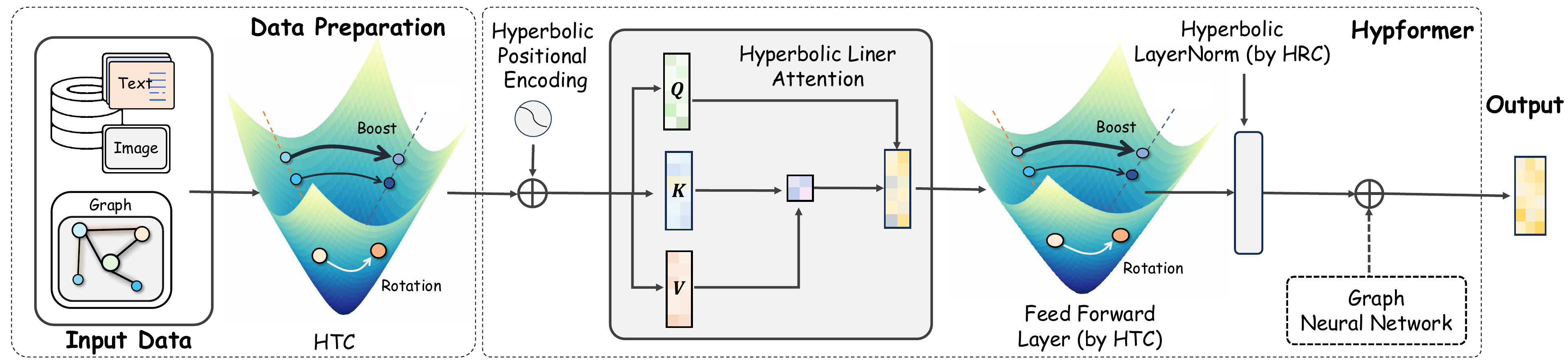}
\caption{Framework of \method. Input data (text, images, graphs) are projected onto the Lorentz model, then transformed via Hyperbolic Linear Transformation (HTC). The result passes through the hyperbolic linear attention block with positional encoding, followed by a Feedforward layer (built by HTC) and LayerNorm (built by HRC). This serves as an encoder which can optionally incorporate a GNN.
For classification tasks in this study, the decoder is the fully connected layer. Dropout, activation, and residual connections are omitted for brevity.
}
\label{fig: framework}
\end{figure*}

\subsection{Hyperbolic Linear Attention}
\label{sec:linear_attention}
 In hyperbolic space, the traditional way of calculating self-attention is quadratic time complexity, which hinders scalability. Therefore, we defined a linear attention through HTC and HRC modules.

Specifically, given the $N$ input token feature with dimension $d$, $\mathbf{X}\in \mathbb{L}^{N\times d, \kappa_1}$ in the Lorentz model with transformation matrix $\mathbf{W}^Q, \mathbf{W}^K, \mathbf{W}^V \in \mathbb{R}^{(d+1)\times d'}$, we first transform it to $\mathcal{Q}, \mathcal{K}$ and $\mathcal{V}$, that is
\begin{equation}
\begin{aligned}
\mathcal{Q} &= \text{HTC}(\mathbf{X}; f_t, \mathbf{W}^Q,\kappa_1, \kappa_2), \\
\mathcal{K} &= \text{HTC}(\mathbf{X}; f_t, \mathbf{W}^K, \kappa_1, \kappa_2), \\
\mathcal{V} &= \text{HTC}(\mathbf{X}; f_t, \mathbf{W}^V,\kappa_1, \kappa_2), \\
\end{aligned}
\end{equation}
where $\mathcal{Q}, \mathcal{K}\text{ and }\mathcal{V} \in \mathbb{L}^{N\times d', \kappa_2}$
Given that the subsequent pairwise similarity computation and aggregation essentially constitute a weighted sum, and their calculation does not involve transformations on the time-like dimension, we adopted the idea of HRC to achieve this. Specifically, we first slice the values of the space-like dimension,
\begin{equation}
    \mathcal{Q}_s, \mathcal{K}_s, \mathcal{V}_s = \phi{(\mathcal{Q}_{[1:]})}, \phi{(\mathcal{K}_{[1:]})}, \phi{(\mathcal{V}_{[1:]})}.
    \label{equ:lorentz_space_like_slicing}
\end{equation}
To achieve linearity, we alter the computation sequence, i.e., transitioning from $(\mathcal{Q}^T\mathcal{K})\mathcal{V}$ to $\mathcal{Q} (\mathcal{K}^T\mathcal{V})$, inspired by~\cite{han2023flatten}.
Our innovation lies in defining this operation in space-like dimensions and recalibrating the time-like value to respect the Lorentz constraint, 
\begin{equation}
    \mathcal{Z}_{s} = \frac{\mathcal{Q}_s(\mathcal{K}_s^T\mathcal{V}_s)}{\mathcal{Q}_s(\mathcal{K}_s^T\mathbf{1})}.
    \label{equ:lorentz_linear_attention}
\end{equation}
Before recalibrating, we incorporate the following residual connection:
\begin{equation}
    \mathcal{\tilde{Z}}_{s} = \mathcal{{Z}}_{s} + \psi(\mathcal{V}_s),
\label{equ:lorentz_rediual_connection}
\end{equation}
and then do the time-like calibration and concatenation,
\begin{equation}
\centering
\begin{aligned}
& \mathcal{Z}_t = \sqrt{\frac{\kappa_2}{\kappa_3}\|\mathcal{\tilde{Z}}_{{s}} \|^2-1/\kappa_3}, &\text{(Time-like~Calibration)}\\
&\mathcal{Z} = \left(\mathcal{Z}_t,  \sqrt{\frac{\kappa_2}{\kappa_3}} \mathcal{\tilde{Z}}_{\text{s}}\right). &\text{(Re-concatenation)}
\label{equ:linear_attention_output}
\end{aligned}
\end{equation}
In Equation (\ref{equ:lorentz_space_like_slicing}, \ref{equ:lorentz_linear_attention}, \ref{equ:lorentz_rediual_connection}), $\mathbf{1}$ denotes an all "1" vector, $\psi$ is a linear layer and $\phi$ signifies the functions employed to enhance the focus of the linear attention, i.e.,
\begin{equation}
\begin{aligned}
\phi(\mathbf{\tilde{e}}) = \frac{\|\mathbf{\tilde{e}}\|}{\|\mathbf{\tilde{e}}^{p}\|}\mathbf{\tilde{e}}^{p}, \text{where }\tilde{\mathbf{e}}_i = \text{ReLU}(\mathbf{e}) / t, 
\label{equ:linear_focused}
\end{aligned}  
\end{equation}
where $\mathbf{e}\in \mathbb{L}^{d',\kappa_2}$ represents the transpose of row in $\mathcal{Q}, \mathcal{K}\text{ and }\mathcal{V}$. 
In this case, $t$ represents a scaling factor, which we set as a trainable parameter in the experiments. The focused strategy is inspired by the work in~\cite{han2023flatten}. A $p>1$ sharpens the paired points, i.e., it enhances the similarity within each group while diminishing the similarity between the groups. Conversely, a $p<1$ has the opposite effect.

This linear attention approach allows us to handle large datasets and long sequences more efficiently while respecting the properties of the Lorentz model. 

\subsection{Hyperbolic Positional Encoding}
Positional encoding in a Transformer model is instrumental in preserving the sequence of input tokens. In what follows, we introduce a relative positional encoding with a trainable model inspired by~\cite{vaswani2017attention,law2019lorentzian}.
\begin{equation}
    \mathbf{\tilde{x}} = \frac{\mathbf{x} + \epsilon\cdot\mathbf{p}}{\sqrt{|\kappa\|\mathbf{x} + \epsilon\cdot\mathbf{p}\|_\mathcal{L}}|}.
\end{equation}
Here, $\mathbf{p} := \mathrm{HTC(\mathbf{x})}$ functions as a Lorentz position vector, and $\epsilon$ specifies the magnitude of $\mathbf{p}$ and we use $1$ in our experiments. This definition calculates the midpoint between $\mathbf{x}$ and $\epsilon\cdot\mathbf{p}$, with respect to the Lorentz constraint. We add the positional encoding before the linear transformation in the self-attention block.
We reserve the exploration of more advanced positional encoding for future works.

\subsection{Hyperbolic LayerNorm, Dropout, Activation, and Concatenation}
\textit{LayerNorm, Dropout, Activation, and Concatenation} are fundamental components of the Transformer architecture. For these operations, we employ HRC in our definitions. This choice is motivated by the fact that these functions are performed within the same reference system and do not involve a time-like dimension. Consequently, we define our operations as follows\footnote{We also include BatchNorm for reference.}:
\begin{equation}
    \begin{aligned}
        \mathrm{HypLayerNorm}(\mathbf{X}) &= \text{HRC}(\mathbf{X},f_{LayerNorm}), \\
        \mathrm{HypBatchNorm}(\mathbf{X}) &= \text{HRC}(\mathbf{X},f_{BatchNorm}), \\
        \mathrm{HypDropout}(\mathbf{X}) &= \text{HRC}(\mathbf{X},f_{Dropout}), \\
        \mathrm{HypActivation}(\mathbf{X}) &= \text{HRC}(\mathbf{X},f_\sigma), \\
        \mathrm{HypConcatnation}(\mathbf{X}) &= \text{HRC}((\mathbf{X}_i, \mathbf{X}_j),f_{concatenation}), \\
    \end{aligned}
\end{equation}
where $f_{Dropout}$, $f_{LayerNorm}$, $f_{BatchNorm}$, and $f_\sigma$ as well as $f_{concatenation}$ represent traditional Euclidean \textit{Dropout}, \textit{LayerNorm}, and \textit{Activation} functions, respectively.
In general, we define $\kappa$ as unchanged before and after the HRC.
In the actual implementation process, for two operations that appear consecutively, such as $f_1 = f_{Dropout}$ and $f_2 = f_{ReLU}$, we merge them into $f = f_1\circ f_2$ for computational efficiency.

\subsection{Overall Architecture}
The framework of \method is shown in Figure~\ref{fig: framework}, it can accept a variety of data types, such as text, images, and graphs.
During the data preparation phase, the input data is mapped to the Lorentz model using an exponential map\footnote{This step is necessary since most data are built from Euclidean space.}. This mapped embedding is then transformed using a HTC layer. In the encoder part of \method, the transformed data is processed through a hyperbolic linear attention block with hyperbolic position encoding. This is followed by the Feedforward layer implemented by HTC, and  LayerNorm layer built by HRC. 
For graph-based inputs, we incorporate the graph neural networks and adopt the parallel paradigm~\cite{min2022transformer} for Transformer and GNN encoder to form a graph Transformer model. 
The processed data is then forwarded to the decoder. 
The decoder can either be the similar structure of encoder, hyperbolic multinomial logistic regression (HypMLR)~\cite{HNN,HNN++} or a tailored design, we leave it in future exploration.
In this research, the decoder is a fully connected layer used for classification tasks.

\begin{table}[t]
\caption{Testing results (ROC-AUC for ogbn-proteins and Accuracy for other datasets) on large-scale node property prediction benchmarks. OOM denotes out of memory during training or testing, and OOT indicates the model could not complete within the allocated time budget. The best and second-best results are highlighted in red bold and underlined, respectively.}
\resizebox{0.45\textwidth}{!}{%
\begin{tabular}{lc|c|c|c}
\toprule
Method                                       & ogbn-proteins                                                  & Amazon2m                                                       & ogbn-arxiv                                                     & Papers100M                         \\ 
\#Nodes                                     & $132,534$                                                      & $2,449,029$                                                    & $169,343$                                                      & $111,059,956$                      \\
\#Edges                                     & $39,561,252$                                                   & $61,859,140$                                                   & $1,166,243$                                                    & $1,615,685,872$                    \\ \midrule\midrule
MLP                                          & $72.0 \pm 0.5$                                                 & $63.5 \pm 0.1$                                                 & $55.5 \pm 0.2$                                                 & $47.2 \pm 0.3$                   \\
GCN~\cite{kipf2016semi}                                          & $72.5 \pm 0.4$                                                 & $83.9 \pm 0.1$                                                 & $71.7 \pm 0.3$                                                 & OOM                                \\
SGC~\cite{wu2019simplifying}                                          & $70.3 \pm 0.2$                                                 & $81.2 \pm 0.1$                                                 & $67.8 \pm 0.3$                                                 & $63.3 \pm 0.2$                   \\
GCN-NSampler                                 & $73.5 \pm 1.3$                                                 & $83.8 \pm 0.4$                                                 & $68.5 \pm 0.2$                                                 & $62.0 \pm 0.3$                   \\
GAT-NSampler                                 & $74.6 \pm 1.2$                                                 & $85.2 \pm 0.3$                                                 & $67.6 \pm 0.2$                                                 & $63.5 \pm 0.4$                   \\
SIGN~\cite{frasca2020sign}                                         & $71.2 \pm 0.5$                                                 & $81.0 \pm 0.3$                                                 & $70.3 \pm 0.3$                                                 & ${65.1} \pm 0.1$          \\
GraphFormer~\cite{ying2021do}                                  & OOM                                                            & OOM                                                            & OOM                                                            & OOM                                \\
GraphTrans~\cite{wu2021representing}                                   & OOM                                                            & OOM                                                            & OOM                                                            & OOM                                \\
GraphGPS~\cite{rampavsek2022recipe}                                     & OOM                                                            & OOM                                                            & OOM                                                            & OOM                                \\
HAN~\cite{gulcehre2019hyperbolicAT}                                     & OOM                                                            & OOM                                                            & OOM                                                            & OOM                                \\
HNN++~\cite{HNN++}                                     & OOM                                                            & OOM & OOM                                                            & OOM \\
F-HNN~\cite{chen2021fully}                                     & OOM                                                            & OOM & OOM                                                            & OOM \\
NodeFormer~\cite{wu2022nodeformer}                                   & ${77.5} \pm {1.2}$                                             & ${87.9} \pm {0.2}$                                             & $59.9 \pm 0.4$                                                 & OOT                                \\
SGFormer~\cite{wu2023simplifying}                                     & {$\underline{79.5 \pm 0.3}$} & 
{$\underline{89.1 \pm 0.1}$} & {$\underline{72.4\pm 0.3}$} & 
{$\underline{65.8 \pm 0.5}$} 
\\ \midrule
\rowcolor{gray!20}Hypformer & \darkred{$\textbf{80.4}\pm \textbf{0.5}$}   & \darkred{$\textbf{89.4}\pm \textbf{0.3}$}   & \darkred{$\textbf{73.2}\pm \textbf{0.2}$}   & $\darkred{\mathbf{66.1}\pm \textbf{0.4}}$                                  \\ \bottomrule
\end{tabular}%
}
\vspace{-10pt}
\label{tab:large_graph}
\end{table}

\textbf{Time complexity.} 
In the proposed \method, the linear attention module is the main computational bottleneck. The complexity comes from two key operations. In Equation~(\ref{equ:lorentz_linear_attention}), we perform a space-like inner product computation of $\mathcal{K}^T$ and $\mathcal{V}$ within the Lorentz model, which incurs a complexity of $\mathcal{O}(d'^2 N)$. Following this, we calculate the inner product of these results with $\mathcal{Q}$, which also has a complexity of $\mathcal{O}(d'^2 N)$. Given that $d'<<N$, the total computational complexity of our method is $\mathcal{O}(N)$.
When dealing with graph inputs, the computational complexity of a GNN model is typical $\mathcal{O}(N + E)$, where $E$ represents the number of edges. Owing to the typical sparsity of graphs (i.e., $E << N^2$), the proposed method can scale linearly with respect to the number of nodes in a graph. This design make \method operate on graphs with billion-level nodes.
\section{Experiments}
\label{sec:experiments}
In this work, we propose a novel hyperbolic Transformer with linear complexity, which is especially well-suited for processing graph-structured data. Graphs often exhibit intricate topological and hierarchical relationships, making them an ideal testbed for evaluating the effectiveness of our proposed hyperbolic Transformer. As such, we primarily focus on comparing our model's performance with other state-of-the-art graph models. 

\subsection{Experiments on Large Graphs}

\textbf{Experimental Settings.} 
We first evaluate \method on diverse large-scale graphs for node classification, with node counts ranging from millions to billions, including ogbn-arxiv, ogbn-protein, and Papers100M (for dataset details, see Appendix~\ref{appendix:data_processing_large_dataset}). To our knowledge, this represents the first application of hyperbolic or non-Euclidean transformations to graphs of this scale. 
Our comparative analysis focuses on state-of-the-art Euclidean GNNs and graph Transformers. We evaluate \method against a spectrum of baselines, including MLP, GCN~\cite{gcn2017}, SGC~\cite{wu2019simplifying}), advanced GNN variants (SIGN~\cite{frasca2020sign}, GCN-NSampler, GAT-NSampler), recent graph Transformer architectures (GraphFormer~\cite{ying2021do}, GraphTrans~\cite{wu2021representing}, GraphGPS~\cite{rampavsek2022recipe}, NodeFormer~\cite{wu2022nodeformer}, SGFormer~\cite{wu2023simplifying}) and hyperbolic models HAN~\cite{gulcehre2019hyperbolicAT}, HNN++~\cite{HNN++} and F-HNN~\cite{chen2021fully}.

\textbf{Experimental Findings}. 
Table~\ref{tab:large_graph} summarizes the results of our experiments. \method consistently outperforms other models across various large-scale graph datasets, demonstrating substantial improvements. It is worth noting that models, such as GraphFormer~\cite{ying2021do}, GraphTrans~\cite{wu2021representing}, and GraphGPS~\cite{rampavsek2022recipe}, HAN~\cite{gulcehre2019hyperbolicAT}, HNN++~\cite{HNN++} and F-HNN~\cite{chen2021fully}, have difficulty operating effectively on large-scale graph data. 
In addition, our method significantly outperforms the recent approaches such as, SGFormer and NodeFormer across all tested scenarios, highlighting its superior effectiveness. Importantly, \method exhibits robust scalability, maintaining its performance advantage even on the largest dataset, ogbn-papers100M, where previous Transformer-based models have encountered limitations.

\subsection{Experiments on Small/Medium Graphs}
To complement our large-scale evaluations, we assessed \method on small- and medium-scale graph datasets. This additional testing allows for a more comprehensive comparison against current state-of-the-art models, including GNNs, graph transformers, and hyperbolic approaches that may not scale effectively to larger datasets. By expanding our evaluation scope, we aim to isolate \method's effectiveness in graph learning from its scalability advantages.

\begin{table}[]
\centering
\caption{Testing results (F1-score for {\sc Disease} and Accuracy for other datasets) on small and medium-sized graph benchmarks. The best and second-best results are highlighted in red bold and underlined, respectively.}
\label{tab:medium_graph_nc}
\resizebox{0.45\textwidth}{!}{
\begin{tabular}{l|c|c|c|c|c}
    \toprule
    Models & {\sc Disease} & {\sc Airport} & {\sc Cora} & {\sc Citeseer}& {\sc PubMed} \\ 
    \#Nodes & $1,044$ & $2,665$ & $2,708$ & $3,327$ & $19,717$ \\
    \#Edges & $1,043$ & $2,664$ & $5,429$ & $4,732$ & $88,651$ \\ \midrule\midrule
    GCN~\cite{gcn2017}                 & $69.7\pm0.4$     & $81.4\pm0.6$     & $81.3\pm0.3$     & $71.6 \pm 0.4$   & $78.1\pm0.2$     \\
    GAT~\cite{velivckovic2017graph}                 & $70.4\pm0.4$     & $81.5\pm0.3$     & $83.0\pm0.7$     & $72.5 \pm 1.1$   & $79.0\pm0.3$     \\
    SGC~\cite{wu2019simplifying}                 & $69.1\pm0.6$     & $82.1\pm0.5$     & $80.1 \pm 0.2$   & $71.9 \pm 0.1$   & $78.7 \pm 0.1$   \\ 
    HGNN~\cite{liu2019HGNN}      & $81.3\pm3.5$  & $84.7\pm1.0$  & $77.1\pm0.8$  & $70.0\pm1.0$  & $78.3\pm1.2$  \\
    HGCN~\cite{chami2019hyperbolic}      & $88.2\pm0.7$  & $89.3\pm1.2$  & $76.5\pm0.6$  & $68.0\pm0.6$  & $78.0\pm1.0$  \\
    HGAT~\cite{chami2019hyperbolic}      & $90.3\pm0.6$  & $89.6\pm1.0$  & $77.4\pm0.7$  & $68.6\pm0.3$  & $78.3\pm1.4$  \\ 
    GraphFormer~\cite{ying2021do}         & $75.2 \pm 0.0$ & $88.1 \pm 1.2$ & $60.0 \pm 0.5$ & $61.4 \pm 0.6$ & $73.3 \pm 0.7$ \\
    GraphTrans~\cite{wu2021representing}          & $89.3 \pm 3.2$ & ${{94.3} \pm {0.6}}$ & $77.6 \pm 0.8$ & $65.1 \pm 1.4$ & $77.5 \pm 0.7$ \\
    GraphGPS~\cite{rampavsek2022recipe}            & $\underline{{92.8} \pm {2.7}}$ & $94.5 \pm 0.9$ & $73.0 \pm 1.4$ & $62.0 \pm 1.5$ & $72.8 \pm 1.4$ \\
    FPS-T~\cite{cho2023curve}             & $88.6\pm 0.9$            & $\darkred{\textbf{96.0}\pm\textbf{0.6}}$   & $82.3\pm0.7$   & $70.0\pm0.7$   & $78.5\pm0.6$   \\
    HAN~\cite{gulcehre2019hyperbolicAT}                 & $85.1\pm0.8$     & $92.9\pm0.6$   & $83.1\pm0.5$     & $72.4\pm0.5$     & 
    $79.0\pm0.6$     \\ 
    HNN++~\cite{HNN++}	&$89.5 \pm 0.2$	&$92.3 \pm 0.3$	&$82.8 \pm 0.6$	&$71.5\pm 1.3$	&$79.9 \pm 0.4$ \\
    F-HNN~\cite{chen2021fully}	&$92.3\pm1.1$	&$93.0\pm0.7$	&$81.0\pm0.7$	&$71.2\pm0.4$	&$77.5\pm0.8$ \\
        NodeFormer~\cite{wu2022nodeformer}          &  $75.9\pm 0.9$            & $80.2\pm0.6$             & $82.2 \pm 0.9$   & $\underline{{72.5} \pm {1.1}}$   & $79.9 \pm 1.0$   \\
    SGFormer~\cite{wu2023simplifying}            & $89.0 \pm 3.9$ & $92.9 \pm 0.5$ & ${\underline{{83.2}\pm {0.9}}}$   & $72.2\pm0.3$   & $\underline{{80.0}\pm{0.8}}$   \\ 
    \midrule
    \rowcolor{gray!20} Hypformer &    $\darkred{\textbf{93.0} \pm 0.7}$          &     $\underline{{95.0} \pm {0.5}}$        & $\darkred{\textbf{85.0}\pm0.3}$   &     $\darkred{\textbf{73.3}\pm 0.4}$         &    $\darkred{\textbf{81.3}\pm 0.3}$          \\ \bottomrule
\end{tabular}
}
\vspace{-10pt}
\end{table}

\begin{table*}[]
\centering
\caption{Experimental results on semi-supervised classification on Mini-ImageNet and 20News-Groups where we use k-NN (with different k's) for artificially constructing an input graph.  The best and second-best results are highlighted in red bold and underlined, respectively.}
\label{tab:my-table}
\resizebox{0.8\textwidth}{!}{
\begin{tabular}{@{}l|cccc|cccc@{}}
\toprule
\multirow{2}{*}{Method} & \multicolumn{4}{c|}{Mini-ImageNet}                                         & \multicolumn{4}{c}{20News-Group}                                          \\ \cmidrule(l){2-9} 
                        & $ k = 5 $      & $ k = 10 $     & $ k = 15 $     & $ k = 20 $     & $ k = 5 $      & $ k = 10 $     & $ k = 15 $     & $ k = 20 $     \\ \midrule
GCN~\cite{gcn2017}      & $84.86 \pm 0.42$ & $85.61 \pm 0.40$ & $85.93 \pm 0.59$ & $85.96 \pm 0.66$ & $65.98 \pm 0.68$ & $64.13 \pm 0.88$ & $62.95 \pm 0.70$ & $62.59 \pm 0.62$ \\
GAT~\cite{gat2018}      & $84.70 \pm 0.48$ & $85.24 \pm 0.42$ & $85.41 \pm 0.43$ & $85.37 \pm 0.51$ & $64.06 \pm 0.44$ & $62.51 \pm 0.71$ & $61.38 \pm 0.88$ & $60.80 \pm 0.59$ \\
DropEdge~\cite{rong2019dropedge} & $83.91 \pm 0.24$ & $85.35 \pm 0.44$ & $85.25 \pm 0.63$ & $85.81 \pm 0.65$ & $64.46 \pm 0.43$ & $64.01 \pm 0.42$ & $62.46 \pm 0.51$ & $62.68 \pm 0.71$ \\
IDGL~\cite{chen2020iterative} & $83.63 \pm 0.32$ & $84.41 \pm 0.35$ & $85.50 \pm 0.24$ & $85.66 \pm 0.42$ & $65.09 \pm 1.23$ & $63.41 \pm 1.26$ & $61.57 \pm 0.52$ & $62.21 \pm 0.79$ \\
LDS~\cite{franceschi2019learning} & OOM              & OOM              & OOM              & OOM              & $\textbf{66.15} \pm 0.36$ & $64.70 \pm 1.07$ & $63.51 \pm 0.64$ & $63.51 \pm 1.75$ \\
NodeFormer~\cite{wu2022nodeformer} & $\underline{86.77 \pm 0.45}$ & $\underline{86.74 \pm 0.23}$ & $\underline{86.87 \pm 0.41}$ & ${86.64 \pm 0.42}$ & $\underline{66.01 \pm 1.18}$ & ${65.21 \pm 1.14}$ & ${64.69 \pm 1.31}$ & ${64.55 \pm 0.97}$ \\
SGFormer                & $86.21 \pm 0.66$ & $86.46 \pm 0.61$ & $86.73 \pm 0.84$ & $\underline{86.76 \pm 0.72}$ & $\darkred{\textbf{68.55} \pm \textbf{0.54}}$ & $\underline{67.96 \pm 0.68}$ & $\underline{66.44 \pm 0.87}$ & $\underline{65.46 \pm 0.59}$ \\ 
NodeFormer w/o graph~\cite{wu2023simplifying}              & \multicolumn{4}{c|}{$\underline{87.46 \pm 0.36}$}                                      & \multicolumn{4}{c}{$64.71 \pm 1.33$}  \\
SGFormer w/o graph~\cite{wu2023simplifying}              & \multicolumn{4}{c|}{$87.25 \pm 0.38$}                                      & \multicolumn{4}{c}{$\underline{67.53 \pm 0.43}$}  \\ \midrule
\rowcolor{gray!20} Hypformer & $\darkred{\textbf{87.36} \pm \textbf{0.73}}$ & $\darkred{\textbf{87.30} \pm \textbf{0.65}}$ & $\darkred{\textbf{87.41} \pm \textbf{0.59}}$ & $\darkred{\textbf{87.48} \pm \textbf{0.61}}$ & $\underline{68.21 \pm 0.78}$ & $\darkred{\textbf{68.01} \pm 0.34}$ & $\darkred{\textbf{66.87} \pm 0.30}$ & $\darkred{\textbf{66.74} \pm 0.19}$ \\
\rowcolor{gray!20} Hypformer w/o graph & \multicolumn{4}{c|}{$\darkred{\textbf{87.73} \pm \textbf{0.63}}$}                                      & \multicolumn{4}{c}{$\darkred{\textbf{67.73} \pm 0.23}$} \\ \bottomrule
\end{tabular}%
\vspace{-10pt}
}
\label{table:image_text}
\end{table*}

\textbf{Experimental Settings.}
We conducted experiments on five small/medium-scale graph datasets, adhering closely to the settings used in HGCN works~\cite{chami2019hyperbolic}. 
These datasets included three low-degree hyperbolicity datasets: {\sc citeseer}, {\sc cora}~\cite{sen2008collective}, and {\sc PubMed}~\cite{namata2012query}, as well as two high-degree hyperbolicity datasets: {\sc Airport} and {\sc Disease}. The number of nodes and edges are shown in Table~\ref{tab:medium_graph_nc}. 
For data split and processing, please refer to Appendix~\ref{appendix:data_processing_small_medium_dataset}.

\textbf{Experimental Findings.}
Table 2 showcases all the experimental results\footnote{Missing values indicate that there were no previous experiments conducted and the results could not be reproduced.}. 
Our findings suggest that the proposed method significantly surpasses both standard GNNs and hyperbolic GNN models by a substantial margin. 
Importantly, the method exhibits effectiveness not only in scenarios with hyperbolic datasets (like {\sc Disease}, {\sc Airport}) but also in situations with non-hyperbolic dataset (like {\sc Cora}, {\sc CiteSeer} and {\sc PubMed}). 
The existing hyperbolic GNN model~\cite{chami2019hyperbolic} had a notable deficiency in this non-hyperbolic datasets. However, by introducing a hyperbolic Transformer, we have successfully overcome this problem. 
This thanks to that Transformers possess long-distance learning capabilities. 
However, on datasets such as {\sc Cora}, {\sc Citeseer}, and {\sc PubMed}, the existing graph Transformers cannot perform well. 
The primary reason might be that the Transformer equals a fully linked aggregation, which will introduce substantial noise. 
Nevertheless, our method employs linear-focused attention to solve this issue effectively.

\subsection{Comparisons on Text and Vision Datasets}
Additionally, we apply our model to semi-supervised image and text classification tasks on the Mini-ImageNet and 20News-Groups datasets. We also construct a graph using k-NN (based on input node features) to utilize graph model. These experiments are conducted closely in Nodeformer. More comprehensive details are provided in Appendix~\ref{appendix:data_processing_text_image_dataset}.
Table~\ref{table:image_text} presents the comparative results for varying $k$ values. Notably, our method outperforms in seven out of eight cases. In contrast, the performance of competing baselines models varying significantly with different k values, while our method demonstrates greater stability.

\section{Analysis}

\begin{figure}[!t]
\centering
\includegraphics[width=0.45\textwidth]{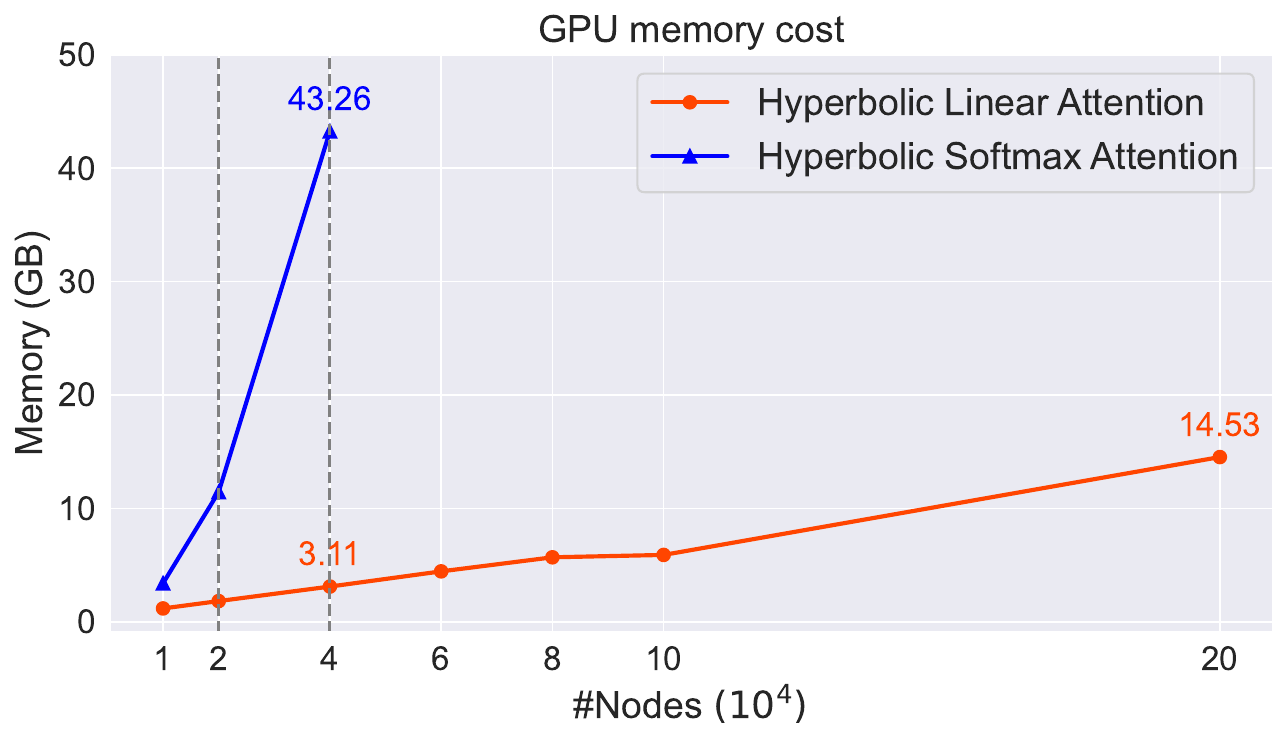}
\caption{Scalability test on an A100 with the proposed linear attention and softmax attention of training time per epoch and GPU memory usage w.r.t. the number of input tokens/nodes.}
\label{fig:Scalability_test}
\end{figure}

\textbf{Scalability of Hypformer.}
We conducted additional tests on the model's scalability regarding the number of nodes in a single batch. 
The Amazon2M dataset was used, and we randomly selected a subset of nodes, with the number of nodes varying from 10K to 200K. We made a comparison between softmax attention defined by Equation~(\ref{equ:self-attention_lorentz}) and linear attention defined by Equation~(\ref{equ:lorentz_linear_attention}), keeping all other parameters the same.
As depicted in Figure~\ref{fig:Scalability_test}, the memory usage of the proposed method exhibits a linear increase with the size of the graph. When the node count exceeds 40K, the softmax attention experiences an out-of-memory (OOM) issue. However, the proposed method continues to function effectively, resulting in a 10X reduction in GPU cost.

\begin{figure}[!t]
\centering
\includegraphics[width=0.23\textwidth]{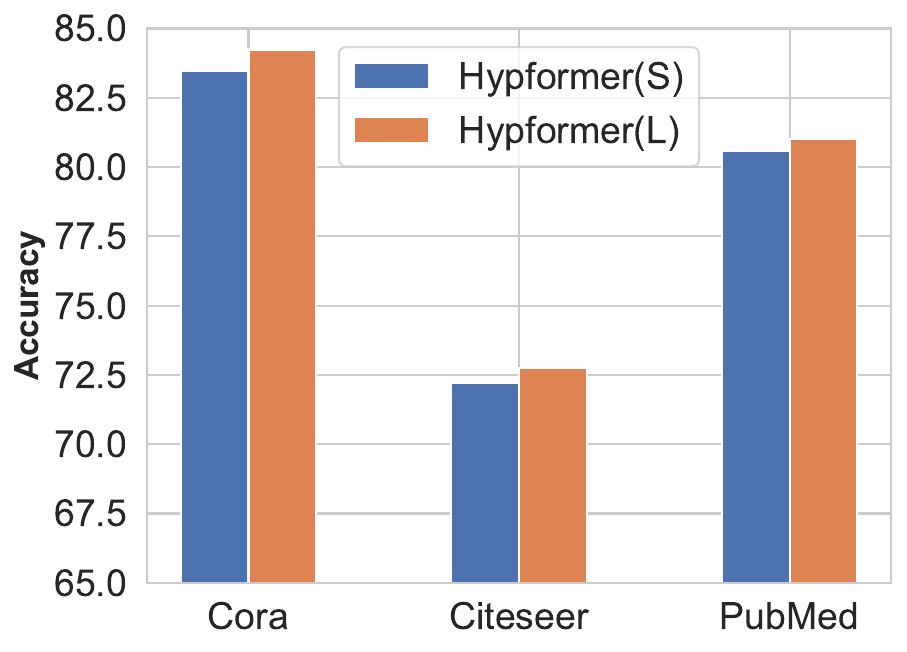}
\includegraphics[width=0.23\textwidth]{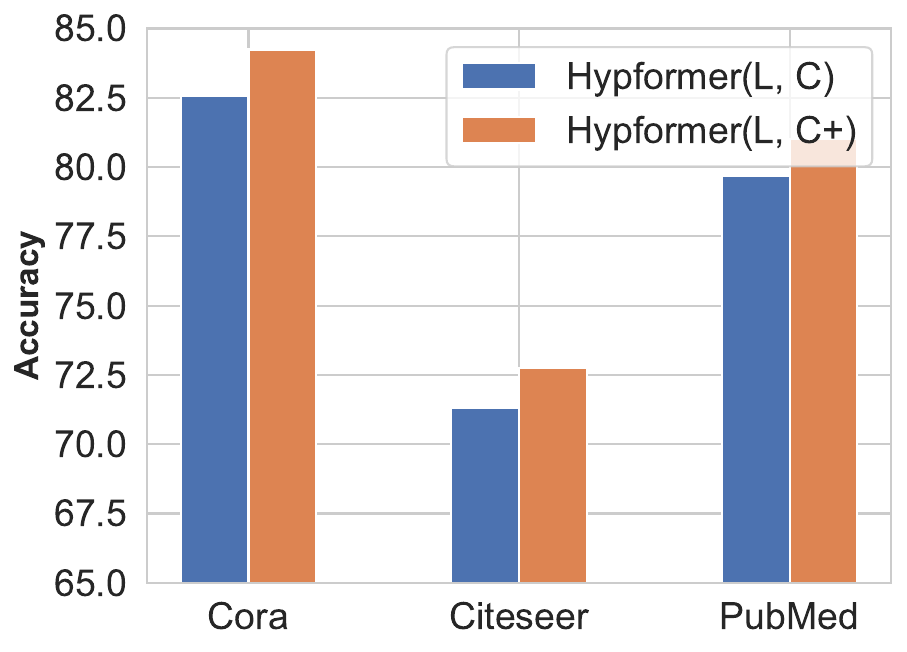}
\caption{Left: Comparison of the proposed linear attention and softmax attention on small/medium datasets. Right: Comparison of unified curvature (Hypformer(L, C)) and varying curvature (Hypformer(L, C+))}
\vspace{-10pt}
\label{fig:parameter_comparsion}
\end{figure}

\textbf{Efficiency and Effectiveness of Hypformer.}
The linear attention designed for \method enhances its efficiency significantly. Table~\ref{table:efficiency_of_hypformer} presents the efficiency of both softmax attention and linear attention within \method.\footnote{To ensure a fair comparison, we have maintained the batch size at a smaller value (40K for Amazon2m and 10K for others) across all tests.}
As indicated in Table~\ref{table:efficiency_of_hypformer}, the proposed linear attention mechanism significantly reduces the training time by half compared to the softmax attention in \method. Furthermore, The left subfigure in Figure~\ref{fig:parameter_comparsion} presents the performance comparison between Hypformer equipped with Softmax attention (Hypformer(S)) and Linear attention (Hypformer(L)). The results demonstrate that both models perform well, with the linear attention exhibiting better accuracy.

\begin{table}[]
\caption{
Efficiency comparison by running time (ms) per epoch between the softmax full and the proposed linear attention in \method on an A100 GPU.
}
\resizebox{0.48\textwidth}{!}{
\begin{tabular}{@{}lcccccc@{}}
\toprule
\multirow{2}{*}{Method} & \multicolumn{2}{c}{ogbn-proteins} & \multicolumn{2}{c}{Amazon2M} & \multicolumn{2}{c}{ogbn-arxiv} \\
                        & Train (ms)           & Test (ms)         & Train (ms)         & Test  (ms)        & Train (ms)         & Test (ms)         \\ \midrule
Hypformer (Softmax)     & 11.9            &  OOM            & 37.38            & OOM              &  7.8             &  OOM            \\
Hypformer (Linear)      & \textbf{5.3}             & \textbf{2.4}           & \textbf{16.32}             & \textbf{2.5}             &\textbf{3}               & \textbf{2.5}             \\
\bottomrule
\end{tabular}
}
\label{table:efficiency_of_hypformer}
\end{table}

\textbf{Effectiveness of Curvature $\kappa$}.
In this work, we propose that both the HTC and HRC basic blocks involve two variable curvatures. 
In our experiment, we set these as trainable parameters. In the right Figure~\ref{fig:parameter_comparsion}, we compare the impact of varying $\kappa$ and fixed curvature on the \method. Experiments show that varying $\kappa$ can always perform better than the unified one.

\textbf{Ablation Study.} To gain a deeper understanding of the proposed Hyperbolic Transformer's effectiveness, we conducted an ablation study on three diverse datasets. We compared the performance of the original Hyperbolic Transformer with two variants: one without the graph component (W/o Graph) and another without the Transformer component (W/o Transformer). The results of this study are presented in Table~\ref{table:ablation_study}.

For the {\sc Cora} dataset, a citation network, removing the graph component leads to a substantial performance drop. This indicates the crucial role of the graph structure in capturing the relationships between nodes in this context. The Transformer component alone (W/o Graph) is insufficient for effectively modeling node interactions. Conversely, removing the Transformer component (W/o Transformer) still yields reasonable performance, highlighting the importance of the graph component for this dataset.
In the case of the ogbn-proteins dataset, which represents a protein-protein interaction network, both the graph and Transformer components contribute significantly to the model's performance. This suggests that the interplay between the graph structure and the Transformer's ability to capture long-range dependencies is essential for accurately modeling the complex interactions in this biological network.
For the 20news dataset, which comprises textual data, the graph is constructed from the original features and may not accurately reflect the true relationships between documents. In this case, the model performs best when the graph component is removed (W/o Graph), indicating that the graph structure might not be as informative for this particular dataset. The Hyperbolic Transformer component alone is sufficient to capture the semantic relationships between documents.
These findings underscore the adaptability of the Hyperbolic Transformer to various datasets and its ability to leverage both graph structure and long-range dependencies when appropriate. 

\begin{table}
\caption{Ablation study}
\resizebox{0.40\textwidth}{!}{
\begin{tabular}{lccc}
\toprule
Dataset & W/o Graph & W/o Transformer & Hypformer \\
\midrule
Cora & $64.6 \pm 0.5$ & $82.8 \pm 0.3$ & $\textbf{85.0} \pm \textbf{0.3}$ \\
ogbn-proteins & $70.0 \pm 0.2$ & $75.9 \pm 0.4$ & $\textbf{80.4} \pm \textbf{0.5}$ \\
Mini-ImageNet & $\textbf{87.7} \pm \textbf{0.6}$ & $85.8 \pm 0.5$ & $87.4 \pm 0.7$ \\
\bottomrule
\vspace{-15pt}
\end{tabular}}
\label{table:ablation_study}
\end{table}

\section{Conclusion}
In this work, we introduce a efficient hyperbolic Transformer, \method. This method operates directly and fully on hyperbolic representations and employs a linear attention mechanism, enabling it to be both scalable and effective. 
Furthermore, this study introduces two basic blocks, HTC and HRC, which are foundational in constructing hyperbolic models. Nonetheless, the research presented is an initial exploration and numerous challenges warrant further investigation. These include the initial determination of a curvature that better reflects the data geometry, the setting of curvature at different levels for \method, and the design of effective decoders for different downstream tasks. We plan to address these issues in our future work.

\section*{Acknowledgements}
We express gratitude to the anonymous reviewers and area chairs for their valuable comments and suggestions.
In this study, Menglin Yang was partly supported by Tony Massini Postdoctoral Fellowship in Data Science from Yale University.
Jiahong Liu and Irwin King were partly supported by the Research Grants Council of the Hong Kong Special Administrative Region, China (CUHK14222922,RGC GRF 2151185).
\bibliographystyle{ACM-Reference-Format}
\balance
\bibliography{references}
\appendix
\section{Exponential and Logarithmic Map}
\label{appendix:exponentail_and_logrithmic_map}
\textbf{Exponential Map}. The exponential map, denoted as $\exp_\mathbf{x}^\kappa: \mathcal{T}_{\mathbf{x}} \mathbb{L}^{n,\kappa}\to \mathbb{L}^{n,\kappa}$, is a function that project any tangent vector $\mathbf{u}$ from the tangent space at point $\mathbf{x}$, $\mathcal{T}_{\mathbf{x}} \mathbb{L}^{n,\kappa}$, to the manifold $\mathbb{L}^{n,\kappa}$, which is given as 
\begin{equation} 
\operatorname{exp}^\kappa_{\mathbf{x}}(\mathbf{u})=\cosh \left(\sqrt{|\kappa|}\|\mathbf{u}\|_{\mathcal{L}}\right) \mathbf{x}+\frac{\sinh \left(\sqrt{|\kappa|}\|\mathbf{u}\|_{\mathcal{L}}\right)}{\sqrt{|\kappa|}\|\mathbf{u}\|_{\mathcal{L}}} \mathbf{u}. 
\end{equation}

\textbf{Logarithmic Map}. 
The logarithmic map $\log _{\mathbf{u}}^\kappa: \mathbb{L}^{n,\kappa} \rightarrow \mathcal{T}_{\mathbf{u}} \mathbb{L}^{n,\kappa}$ plays an opposite role, more specifically, 
\begin{equation}
    \log _\mathbf{u}^\kappa(\mathbf{x})=\frac{\cosh ^{-1}\left(\kappa\langle \mathbf{u}, \mathbf{x}\rangle_{\mathcal{L}}\right)}{\sinh \left(\cosh ^{-1}\left(\kappa\langle \mathbf{u}, \mathbf{x}\rangle_{\mathcal{L}}\right)\right)}\left(\mathbf{x}-\kappa\langle \mathbf{u}, \mathbf{x}\rangle_{\mathcal{L}} \mathbf{u}\right).
\end{equation}

 \textbf{Lorentz Distance}. The Lorentz distance between two points $(\mathbf{x}\in \mathbb{L}^{n,\kappa}, \mathbf{y}\in \mathbb{L}^{n,\kappa})$ is given as:
\begin{equation}
    d_{\mathcal{L}}^\kappa(\mathbf{x}, \mathbf{y})=\frac{1}{\sqrt{|\kappa|}} \cosh ^{-1}\left(\kappa\langle \mathbf{x}, \mathbf{y}\rangle_{\mathcal{L}}\right)
\end{equation}

\section{Proof}
\textbf{Proof of Proposition~\ref{prop:lorentz_constraint}}
\label{appendix:proof_constraint}
\begin{proof}
    Let $\mathbf{L}_\mathbf{x}= \operatorname{LTC}(\mathbf{x}; f_t; \mathbf{W},\kappa_a, \kappa_b)$ and $\langle \mathbf{L}_\mathbf{x}, \mathbf{L}_\mathbf{x} \rangle_\mathcal{L} = 1/\kappa_b$ holds. Besides, $f_t(\mathbf{x};\mathbf{W}): \mathbb{R}^{d_a+1}\to \mathbb{R}^{d_b}$. With the time-like dimension re-calibration and concatenation, $\mathbf{L}_\mathbf{x}\in \mathbb{R}^{d_b + 1}.$
    Therefore $\mathbf{L}_\mathbf{x}\in \mathbb{L}^{d_b,\kappa_b}$
        \label{prop:lorentz_preserve}
\end{proof}

\textbf{Proof of Proposition~\ref{prop:lorentz_preserving_text}}
\begin{proof}
First, let $$\mathbf{z} = \left(\sqrt{\| f(\mathbf{x}; \mathbf{W}) \|_2^2-1/\kappa_1}, f(\mathbf{x}; \mathbf{W})\right)$$ and then $$\mathrm{HTC}(x,W, \kappa_1, \kappa_2)= \sqrt{\frac{\kappa_1}{\kappa_2}}\cdot \mathbf{z}.$$ 
We know $\mathbf{z}\in \mathbb{L}^{\kappa_1}$ and $\mathbf{z}'= \mathrm{HTC}(x,W, \kappa_1, \kappa_2)\in \mathbb{L}^{\kappa_2}$.
Consider the distance between any pair of points in $\mathbb{L}^{\kappa_1}$ and $\mathbb{L}^{\kappa_2}$:
\begin{equation}
\begin{aligned}
    d_{\mathcal{L}}^{\kappa_1}\left(\mathbf{z}_i, \mathbf{z}_j\right) &= \sqrt{1/|\kappa_1|}\text{arcosh}\left(\kappa_1\langle \mathbf{z}_i, \mathbf{z}_j \rangle_\mathcal{L}\right),\\
\end{aligned}
\label{equ:distance_k1_ij}
\end{equation}

\begin{equation}
\begin{aligned}
    d_{\mathcal{L}}^{\kappa_1}\left(\mathbf{z}_i, \mathbf{z}_k\right) &= \sqrt{1/|\kappa_1|}\text{arcosh}\left(\kappa_1\langle \mathbf{z}_i, \mathbf{z}_k \rangle_\mathcal{L}\right),\\
\end{aligned}
\label{equ:distance_k1_ik}
\end{equation}

\begin{equation}
\begin{aligned}
    d_{\mathcal{L}}^{\kappa_2}\left(\mathbf{z}_i', \mathbf{z}_j'\right) &= \sqrt{1/|\kappa_2|}\text{arcosh}\left(\kappa_2\langle \mathbf{z}_i', \mathbf{z}_j' \rangle_\mathcal{L}\right),\\
\end{aligned}
\label{equ:distance_k2_ij}
\end{equation}

\begin{equation}
\begin{aligned}
    d_{\mathcal{L}}^{\kappa_2}\left(\mathbf{z}_i', \mathbf{z}_k'\right) &= \sqrt{1/|\kappa_2|}\text{arcosh}\left(\kappa_2\langle \mathbf{z}_i', \mathbf{z}_k' \rangle_\mathcal{L}\right).\\
\end{aligned}
\label{equ:distance_k2_ik}
\end{equation}
\darkred{We aim to prove that if
\begin{equation}
    d_{\mathcal{L}}^{\kappa_1}\left(\mathbf{z}_i, \mathbf{z}_j\right)\geq 
    d_{\mathcal{L}}^{\kappa_1}\left(\mathbf{z}_i, \mathbf{z}_k\right),
    \label{eq:inequality_k1}
\end{equation}
then
\begin{equation}
\begin{aligned}
d_{\mathcal{L}}^{\kappa_2}\left(\mathbf{z}_i', \mathbf{z}_j'\right)\geq d_{\mathcal{L}}^{\kappa_2}\left(\mathbf{z}_i', \mathbf{z}_k'\right).
\end{aligned}
\label{eq:inequality_k2}
\end{equation}}
Let us expand $d_{\mathcal{L}}^{\kappa_2}\left(\mathbf{z}_i', \mathbf{z}_j'\right)$:
\begin{equation}
\begin{aligned}
d_{\mathcal{L}}^{\kappa_2}\left(\mathbf{z}_i', \mathbf{z}_j'\right) &= \sqrt{1/|\kappa_2|}\text{arcosh}\left(\kappa_2\langle \mathbf{z}_i', \mathbf{z}_j' \rangle_\mathcal{L}\right),\\
    &= \sqrt{1/|\kappa_2|}\text{arcosh}\left(\kappa_2\sqrt{\frac{\kappa_1}{\kappa_2}}\cdot\sqrt{\frac{\kappa_1}{\kappa_2}}\langle \mathbf{z}_i, \mathbf{z}_j\rangle_\mathcal{L}\right) \\
    & = \sqrt{1/|\kappa_2|}\text{arcosh}\left(\kappa_1 \langle\mathbf{z}_i, \mathbf{z}_j\rangle_\mathcal{L}\right) \\
    & = \sqrt{\frac{\kappa_1}{\kappa_2}}\left(d_{\mathcal{L}}^{\kappa_1}\left(\mathbf{z}_i, \mathbf{z}_j\right)\right)
\end{aligned}
\label{eq:distance_relation_ij}
\end{equation}
Similarly, we can show that:
\begin{equation}
d_{\mathcal{L}}^{\kappa_2}\left(\mathbf{z}_i', \mathbf{z}_k'\right) = \sqrt{\frac{\kappa_1}{\kappa_2}}d_{\mathcal{L}}^{\kappa_1}\left(\mathbf{z}_i, \mathbf{z}_k\right)
    \label{eq:distance_relation_ik}
\end{equation}
\darkred{Given the inequality in \eqref{eq:inequality_k1}, we can multiply both sides by $\sqrt{\frac{\kappa_1}{\kappa_2}}$ (which is positive):}
\begin{equation}
    \sqrt{\frac{\kappa_1}{\kappa_2}}d_{\mathcal{L}}^{\kappa_1}\left(\mathbf{z}_i, \mathbf{z}_j\right) \geq \sqrt{\frac{\kappa_1}{\kappa_2}}d_{\mathcal{L}}^{\kappa_1}\left(\mathbf{z}_i, \mathbf{z}_k\right)
\end{equation}
Substituting from \eqref{eq:distance_relation_ij} and \eqref{eq:distance_relation_ik}, we obtain:
\begin{equation}
d_{\mathcal{L}}^{\kappa_2}\left(\mathbf{z}_i', \mathbf{z}_j'\right) \geq d_{\mathcal{L}}^{\kappa_2}\left(\mathbf{z}_i', \mathbf{z}_k'\right).
\end{equation}
This proves the desired inequality \eqref{eq:inequality_k2}.
\end{proof}

\section{Data Processing and Experimental Details}
\subsection{Data Processing for Large-graph Data} 
\label{appendix:data_processing_large_dataset}
We employ the public splits offered by OGB~\cite{hu2020open} for ogbn-proteins and ogbn-arxiv datasets.
Additionally, we assess our approach using models on the Amazon2M item co-occurrence network, which comprises 2.45 million nodes and 61.86 million edges. For Amazon2M, we follow the same splits used in recent studies~\cite{wu2022nodeformer,wu2023simplifying}. The largest dataset we employ is ogbn-papers100M, boasting an impressive 0.11 billion nodes and 1.61 billion edges. We also adhere to the publicly available OGB splits for this dataset.

\subsection{Data Processing for Medium-graph Data}
\label{appendix:data_processing_small_medium_dataset}
We used standard splits~\cite{kipf2016semi} for the citation networks. For the Airport and Disease datasets, the train/val/test splits were 70\%/15\%/15\% and 30\%/10\%/60\%, respectively, which is the same as~\cite{chami2019hyperbolic}. 
We report the results of five runs on the node classification task. For {\sc Disease} and {\sc Airport}, which are imbalanced, we report the F1-score. For the other datasets, we report the accuracy.
\textbf{Baselines.} 
For the baselines, we compare \method against the basic GNN models, including GCN~\cite{kipf2016semi}, GAT~\cite{velivckovic2017graph} and SGC~\cite{wu2019simplifying}. 
For Hyperbolic GNN models, we utilized HGCN \cite{chami2019hyperbolic}, LGCN~\cite{zhang2021lorentzian} and HGNN~\cite{feng2019hypergraph} as the competitors. 
We also compared with state-of-the-art Euclidean graph Transformers models viz. Graphomer \cite{ying2021do}, GraphTrans \cite{wu2021representing}, GraphGPS \cite{rampavsek2022recipe}, NodeFormer \cite{wu2022nodeformer} and SGFormer \cite{wu2023simplifying}. Graphormer suggested the incorporation of edge connectivities into the model by employing shortest-path distances to bias the attention mechanism. GraphTrans introduced a permutation-invariant Transformer module combined with a GNN module. While, Nodeformer, GraphGPS and SGFormer each introduced linear attention mechanisms. Specifically, Nodeformer employed a kernelized Gumbel-Softmax, while GraphGPS seperated the local real-edge aggregation and the fully-connected Transformer to achieve this complexity. Besides, we also compared with non-Euclidean transformers, i.e., the HAN~\cite{gulcehre2019hyperbolicAT} and FPS-T~\cite{cho2023curve}.

\subsection{Data Processing for Text and Image Data}

\label{appendix:data_processing_text_image_dataset}

We tested our model on two datasets without a graph structure: 20News-Groups and Mini-ImageNet. 
For our experiment, we selected 30 classes from the dataset, each with 600 images with 128 features extracted by a CNN. These settings closely follow the Nodeformer~\cite{wu2022nodeformer}. For each dataset, we randomly allocate instances into training, validation, and testing sets, comprising 50\%, 25\%, and 25\% of the data, respectively. {Following existing works~\cite{semivn,vgatm,netdtm}}, we also construct a graph using k-NN (based on input node features) to facilitate the message passing of GNN and the graph transformer.
All the datasets we used in the experiment were directly sourced, except for Mini-ImageNet, for which we extracted the features ourselves. Following the approach of ~\cite{wu2022nodeformer}, we computed node embeddings using a CNN model with four convolutional layers followed by a fully connected layer, resulting in a 128-dimensional embedding. These 128-dimensional outputs are then used as the features of the nodes (images) for subsequent tasks based on Graph Neural Networks (GNNs).

\subsection{Implementation Details}
In the reported results, we mainly refer to findings from several relevant works for the baseline comparisons~\cite{wu2019simplifying,chami2019hyperbolic,yang2023kappahgcn}. 
For the most relevant studies, such as SGFormer, other Graph Transformers, we reproduce results using identical experimental settings to ensure a fair comparison. 
It is important to note that the results of SGFormer cannot be fully reproduced due to errors in its official code implementation. 
To maintain the integrity of our analysis, we report the performance of SGFormer based on the available information while acknowledging the discrepancy caused by the implementation issues.
Our experimental setup for \method mainly follows the configurations used in SGformer. 
Additionally, we performed parameter tuning for the input curvature and output curvature, exploring values within [1.0, 2.0, 3.0]. This is grounded in our hypothesis that the input attributes and hidden states belong to different curvature spaces. While a more detailed curvature setting could be employed, we leave this for future exploration. Furthermore, we conducted a parameter search for $p$ in Equation~(\ref{equ:linear_focused}) within [1.0, 2.0, 3.0]. Regarding the decoder, we created Euclidean and hyperbolic classifiers for experiments, with the Euclidean classifier performing better in most cases.

\end{document}